\documentclass[final,3p, 10pt]{elsarticle}

\usepackage{graphics}
\usepackage{graphicx}
\usepackage{amsfonts}
\usepackage{amssymb}
\usepackage{amsmath}
\usepackage{dsfont}
\usepackage{color}
\usepackage{natbib,hyperref}
\usepackage{subcaption}
\usepackage{epsfig}
\usepackage{tikz}
\usetikzlibrary{bayesnet}
\usepackage{float}

\usepackage{color}

\usepackage[ruled]{algorithm2e}
\usepackage{algorithmic}

\definecolor{violet}{cmyk}{.55,.25,0,.1}
\definecolor{burntorange}{cmyk}{0,0.52,1,0}

\def\viol{violet!90}
\def\oran{orange!30}
\tikzstyle{output}=[rectangle, draw, violet, rounded corners, thin, bottom color=\viol, top color=white,text=black]
\tikzstyle{infer}=[circle, draw, burntorange, rounded corners, thin, bottom color=\oran, top color=white,text=black]

\newcommand{\by}{\mathbf{y}}

\newcommand{\bh}{\mathbf{h}}
\newcommand{\bu}{\mathbf{u}}
\newcommand{\bv}{\mathbf{v}}
\newcommand{\ba}{\mathbf{a}}

\newcommand{\bc}{\mathbf{c}}
\newcommand{\bw}{\mathbf{w}}

\newcommand{\bm}{\mathbf{m}}

\newcommand{\bb}{\mathbf{b}}

\newcommand{\bL}{\mathbf{L}}
\newcommand{\bH}{\mathbf{H}}

\newcommand{\bM}{\mathbf{M}}

\newcommand{\bell}{\boldsymbol{\ell}}

\newcommand{\bepsilon}{\boldsymbol{\epsilon}}

\newcommand{\bsigma}{\boldsymbol{\sigma}}
\newcommand{\bvsigma}{\boldsymbol{\varsigma}}
\newcommand{\brho}{\boldsymbol{\rho}}
\newcommand{\bxi}{\boldsymbol{\xi}}
\newcommand{\bXi}{\boldsymbol{\Xi}}

\newcommand{\btheta}{\boldsymbol{\theta}}
\newcommand{\balpha}{\boldsymbol{\alpha}}
\newcommand{\bbeta}{\boldsymbol{\beta}}

\newcommand{\bfeta}{\boldsymbol{\eta}}

\newcommand{\bzeta}{\boldsymbol{\zeta}}
\newcommand{\blambda}{\boldsymbol{\lambda}}

\newcommand{\bPsi}{\boldsymbol{\Psi}}

\newcommand{\bpi}{\boldsymbol{\pi}}

\newcommand{\bTheta}{\boldsymbol{\Theta}}
\newcommand{\biota}{\boldsymbol{\iota}}

\newcommand{\calB}{\mathcal{B}}
\newcommand{\calG}{\mathcal{G}}
\newcommand{\calL}{\mathcal{L}}
\newcommand{\calD}{\mathcal{D}}
\newcommand{\calN}{\mathcal{N}}

\newcommand{\calF}{\mathcal{F}}

\newcommand{\calI}{\mathcal{I}}

\newcommand{\calU}{\mathcal{U}}
\newcommand{\calX}{\mathcal{X}}

\newcommand{\calH}{\mathcal{H}}

\newcommand{\sfB}{\mathsf{B}}

\newcommand{\ww}{\mathrm{w}}

\newcommand{\trace}[1]{\operatorname{Tr}\left[#1\right]}

\newcommand{\KL}{\operatorname{KL}}
\newcommand{\diag}[1]{\operatorname{diag}\left(#1\right)}

\newcommand{\R}{\mathbb{R}}
\newcommand{\N}{\mathbb{N}}
\newcommand{\E}{\mathbb{E}}

\begin{document}

\begin{frontmatter}

\title{Sparse Polynomial Chaos expansions using Variational Relevance Vector Machines}
\date{}

\author[label1]{Panagiotis Tsilifis\corref{cor1}}
\ead{panagiotis.tsilifis@epfl.ch}

\author[label2]{Iason Papaioannou}
\ead{iason.papaioannou@tum.de}

\author[label2]{Daniel Straub}
\ead{straub@tum.de}

\author[label1]{Fabio Nobile}
\ead{fabio.nobile@epfl.ch}

\cortext[cor1]{Corresponding author}
\address[label1]{CSQI, Institute of Mathematics, \'{E}cole Polytechnique F\'{e}d\'{e}rale de Lausanne, Lausanne, CH-1015, Switzerland}
\address[label2]{Engineering Risk Analysis Group, Technische Universit\"{a}t Munchen, Arcisstr. 21, M\"{u}nchen, Germany}



\begin{abstract}

The challenges for non-intrusive methods for Polynomial Chaos modeling lie in the computational efficiency and accuracy under a limited number of model simulations. These challenges can be addressed by enforcing sparsity in the series representation through retaining only the most important basis terms. In this work, we present a novel sparse Bayesian learning technique for obtaining sparse Polynomial Chaos expansions which is based on a Relevance Vector Machine model and is trained using Variational Inference. The methodology shows great potential in high-dimensional data-driven settings using relatively few data points and achieves user-controlled sparse levels that are comparable to other methods such as compressive sensing. The proposed approach is illustrated on two numerical examples, a synthetic response function that is explored for validation purposes and a low-carbon steel plate with random Young's modulus and random loading, which is modelled by stochastic finite element with $38$ input random variables.

\end{abstract}

\end{frontmatter}

\section{Introduction}
\label{sec:intro}

The use of probabilistic approaches in engineering systems for risk assessment and reliability analysis has been established throughout the last few decades. In this context, the rapidly increasing availability of computational resources has resulted in a steady transition from Uncertainty Quantification (UQ) problems with moderate amounts of data to problems with massive data. Therefore, developing more complex and elegant methodologies to analyze this data is of paramount importance. Statistical machine learning research attempts to address several challenges related to the data complexity and to reveal hidden structures and dependencies in high-dimensional and nonlinear models \cite{bishop2006, murphy2012}. This can be achieved for example with the use of \emph{probabilistic graphical models} \cite{koller_friedman}, \emph{kernel methods (i.e. Gaussian Processes)} \cite{williams_rasmussen, bilionis_GP, raissi_perdikaris_GP} or \emph{sparse kernel machines} \cite{cortes_vapnik, tipping}, which allow one to visualise such structures and to quantify specific dependencies using posterior inference algorithms. However, expensive computer codes result in paucity of data, therefore, the challenge of training high-dimensional surrogate models or performing regression tasks becomes again cumbersome. 

This work focuses on investigating uncertainty propagation in physical systems where the system response depends on a large number of uncertain inputs through a computational model that we seek to replace with a computationally inexpensive emulator. Particularly, we are interested in regression models that support capabilities for sparse representation through sparse basis expansions. Sparse regression techniques such as Compressive Sensing (CS) \cite{baraniuk, ji}, LASSO \cite{tibshirani}, the Support Vector Machine (SVM) \cite{cristianini} or the Relevance Vector Machine (RVM) \cite{tipping} have become popular in the context of image processing \cite{zhou_nips, zhou_ieee}, natural language processing \cite{chaspari, chaspari_tsil} and only lately in UQ applications \cite{peng_doostan, hampton_doostan, adcock, x_huan, bilionis_RVM, rauhut}.
Therein, sparse regression is used to cope with the presence of expensive computer codes that poses strict limitations on the number of experiment repetitions and therefore on the size of the data that are used to train regression models. A common characteristic of regression methods with sparse basis expansions is that they all expand linearly some physical output in a series of basis functions or dictionary atoms, which are typically taken to be kernel functions or orthogonal polynomials.

Throughout this manuscript, we take the basis functions to consist of orthogonal polynomials that are functions of the physical input parameters on which a probability measure is imposed. The resulting representation, commonly known as Polynomial Chaos expansion (PCE), is a well-established spectral representation technique, that was introduced first by Wiener \cite{wiener} and was later applied in engineering problems within a finite-element context \cite{ghanem_spanos, Ghanem:1999}. The key property of the PCE, in its original form, is that the basis functions consisting of multivariate Hermite polynomials of the Gaussian input parameters, span the space of square integrable random variables and random processes \cite{cameron, janson}. The generalized Polynomial Chaos \cite{xiu_karniadakis, soize_physical} further allows to choose the type of polynomials depending on the probability measure of the input variables such that orthogonality is guaranteed and series truncation leads to best approximation in the mean squared sense. Although it was initially applied in a Galerkin-projection setting \cite{xiu_book, xiu_karniadakis}, non-intrusive approaches were also developed, in order to cope with black-box simulation models and more complex forward propagation problems.
Non-intrusive techniques using pseudo-spectral numerical integration \cite{debusschere, reagan}, interpolation \cite{babuska, nobile, xiu_colloc, xiu_hesthaven} or least-squares regression \cite{berveiller2006stochastic} work well in moderate dimensions and low polynomial order settings; however, these techniques are particularly challenged when the number of points and therefore, the number of required forward simulations increases rapidly as a function of the dimension.

In order to overcome this curse of dimensionality, several alternative methods for computing the chaos coefficients have been proposed in the literature, such as alternating least-squares regression \cite{doostan_iacc}, least-angle regression \cite{blatman}, $\ell_1$-minimization \cite{adcock, peng_doostan, yang_l1}, Bayesian compressive sensing \cite{sargsyan_IJUQ}, maximum likelihood \cite{desceliers} and adaptive least squares \cite{bachmayr_i, bachmayr_ii}, targeting primarily the reduction of the chaos representation by enforcing some notion of sparsity. In a different spirit, dimensionality reduction techiques have been developed within the Polynomial Chaos context, based on the idea of applying rotations on the Gaussian input \cite{tipireddy, tsilifis_wiener}. This idea of adapting the input basis, when applicable, enables the efficient estimation of a low dimensional PCE, using standard non-intrusive techniques. Several criteria to choose the rotation matrix, along with estimating the associated coefficients have been explored, including active subspaces \cite{tsilifis_ASME}, compressed sensing \cite{tsilifis_BACS}, partial least squares \cite{papaioannou} and Bayesian inference \cite{tsilifis_geod} and the resulting adapted PCEs have shown great potential in design optimization and Bayesian inference problems \cite{thimmisetty, ghauch}.

Although the Bayesian estimation of chaos coefficients is not new, previous methods \cite{arnst, ghanem_doostan, sargsyan_IJUQ} have no direct benefits in terms of computational savings and the data requirements for obtaining tight posteriors remain significant. This paper explores a new way of computing PCE coefficients using a Bayesian formalism, namely a Variational Inference (VI) technique applied on the above-mentioned RVM. The approach follows the steps of the VI technique used in \cite{tsilifis_geod}, with the key difference that the current setting further employs a sparsity variable that allows dropping basis terms that have no \emph{a-posteriori} influence on the system's output, thus, making it attractive for high-dimensional applications. VI is an approximate Bayesian inference technique that transforms the problem of posterior density exploration to an optimization problem that determines the parameters of a parametric family of densities, such that the Kullback-Leibler divergence \cite{kullback} from the target posterior density is minimized. VI has enjoyed extended popularity in the machine learning community \cite{bishop_tipping, tipping, tipping_RVM, tipping_patent, attias, hoffman}; only recently it has been used in UQ works for Bayesian Inversion \cite{pinski, tsilifis_VI, tsilifis_QCM, franck_koutsou}, uncertainty propagation \cite{chen_bilionis} and for training physics-informed neural nets \cite{zhu_zabaras, yang_paris}. Here VI is used in order to approximate posterior densities of the chaos coefficients, conditioned on data that consists of model input and output points. 
The prior assumptions follow the RVM formalism that introduces a hierarchic prior structure governed by a set of hyperparameters.
RVM has several known benefits over its deterministic counterpart, the support vector machine (SVM), which requires cross validation techniques and poses restrictions on the choice of basis functions (\cite{bishop2006}, $\S 7.2$). Unlike the traditional training procedure in RVM that tunes the hyperparameters by maximizing the marginal likelihood \cite{tipping, bilionis_RVM, sargsyan_IJUQ}, VI approximates the posterior densities by an element of the same exponential family density as the priors. 
The VI optimization problem is solved using a gradient ascent scheme wherein the gradient of the objective function can be evaluated explicitly due to the particular choice of the parametric family.
To the best of our knowledge, this is the first work that attempts an RVM-with-VI training approach in the context of Polynomial Chaos expansions and our method appears to outperform other sparsity-enhanced methodologies such as $\ell_1$-minimization.

This paper is structured as follows: Section \ref{sec:pcRVM} formulates the RVM model using a sparse Polynomial Chaos expansion as the response function. Section \ref{sec:VI} presents the variational inference framework for approximating the posterior densities of the PCE coefficients, the sparsity vector and their hyperparameters. Finally Section \ref{sec:num_ex} illustrates the performance of the methodology with two numerical examples.

\section{Polynomial Chaos Relevance Vector Machine}
\label{sec:pcRVM}

\subsection{Polynomial Chaos expansion}

Let us represent by a mapping $f: \calX\subset  \R^K \to \R$, the result of a deterministic solver which models a physical system that receives input values $\bxi \in \calX$ and returns output $f(\bxi)$. We treat $\bxi$ as an $\R^K$-valued random variable and we denote  with $p_{\bxi}(\bxi)$ its probability density. We also assume that the variance, or equivalently, the second moment of the output quantity of interest (QoI) is finite, that is $\int_{\calX} f(\bxi)^2 p_{\bxi}(\bxi) d\bxi < +\infty$. Naturally $f$ lies in the space of square integrable random variables, denoted with $\calL_2$, that forms a Hilbert space \cite{cameron} with inner product
\begin{equation}
\langle g, h \rangle_{\calL_2} = \int_{\calX} g(\bxi) h(\bxi) p_{\bxi}(\bxi)d\bxi, \ \ g, h \in \calL_2.
\end{equation}
The space  $\calL_2$ is spanned by a basis $\left\{\Psi_{\balpha}(\cdot) : \balpha \in \N^K\right\}$ of $K$-dimensional polynomials that are orthogonal with respect to $\langle\cdot, \cdot \rangle_{\calL_2}$, that is
\begin{equation}
\label{eq:orthpol}
\int_{\calX} \Psi_{\balpha}(\bxi) \Psi_{\bbeta}(\bxi) p_{\bxi}(\bxi)d\bxi = \vert\vert \Psi_{\balpha} \vert\vert^2 \delta_{\balpha\bbeta},
\end{equation}
where $\delta_{\balpha\bbeta}$ is the Kronecker delta, taking the value $1$ for $\alpha_i = \beta_i$, $i = 1, \dots, K$ and $0$ otherwise.
Throughout this work we assume that the vector $\bxi = (\xi_1,\dots,\xi_K)^T$ consists of independent and identically distributed random components (iid), which implies the joint density factorization $p_{\bxi}(\bxi) = \prod_{i=1}^K p_{\xi}(\xi_i)$. 
We note that this assumption is not very restrictive, as it is typically possible to represent dependent inputs as functions of iid random variables through probabilistic transformation techniques \cite{rosenblatt1952remarks}.
In the iid case, the multidimensional basis polynomials are given by  
\begin{equation}
\Psi_{\balpha}(\bxi) = \prod_{i=1}^K \psi_{\alpha_i}(\bxi_i), 
\end{equation}
where $\{\psi_n(\cdot): n\in \N\}$, are univariate polynomials, $n$ denoting the degree, that are constructed to be orthogonal with respect to $p_{\xi}(\cdot)$ \cite{gautschi}. Note that this independence assumption is not required in order to develop the RVM methodology that will follow. For simplicity we also assume that the polynomials are normalized by dividing them by $\vert\vert \Psi_{\balpha} \vert\vert^2$, for all $\balpha$, therefore our basis is in fact orthonormal. The total degree of $\Psi_{\balpha}$ is defined as $\vert\balpha\vert = \sum_{i=1}^K\alpha_i$.

From the above, we have that our QoI $f$ admits a spectral representation
\begin{equation}
\label{eq:pce}
f_{pc}(\bxi) = \sum_{\balpha \in \N^K} f_{\balpha}\Psi_{\balpha}(\bxi), 
\end{equation}
where the square integrability and the orthonormality of the basis functions implies that $\sum_{\vert\balpha\vert = 0}^{\infty} f^2_{\balpha} < +\infty$, thus indicating a decaying rate for the series coefficients. For practical use we typically consider a truncated version of (\ref{eq:pce}), that is
\begin{equation}
\label{eq:pce_tr}
f_{pc}(\bxi) = \sum_{\vert\balpha\vert = 0}^{P} f_{\balpha} \Psi_{\balpha}(\bxi),
\end{equation}
where the summation is over all multi-indices of maximum total degree $P\in \N$ and it is referred to as \emph{Total Degree (TD) truncation}, containing $N_{K, P} = \binom{K + P}{P} = \frac{(K + P)!}{K!P!}$ terms. Other truncation schemes that have been used in the literature include the \emph{$\ell_q$ (LQ) truncation}, where the summation is defined over the set of multi-indices $\balpha$ such that $\vert\vert \balpha \vert\vert_{\ell_q} :=  \left(\sum_{i=1}^K \alpha_i^q\right)^{1/q} \leq P$, for some $q \in (0, 1]$, the \emph{Tensor Product (TP) truncation} that sums over the set of $\balpha$ with $\max_i \{\alpha_i\} \leq P$ and the \emph{Hyperbolic-cross (HC) truncation} that sums over the set of $\balpha$ with $\prod_{i=1}^K(\alpha_i + 1) \leq P+1$ \cite{beck_2011, beck_2014}. For notational convenience we recast the series as 
\begin{equation}
f_{pc}(\bxi) = \sum_{i=1}^{N_{K, P}} \ww_i \Psi_i(\bxi),
\end{equation}
such that there is a one-to-one correspondence between $\{f_{\balpha}, \Psi_{\balpha}\}$ defined in (\ref{eq:pce_tr}) and $\{\ww_i, \Psi_i\}$ respectively. For simplicity, unless otherwise stated, we keep $P$ fixed and we simply write $N_K$ for $N_{K,P}$. 

\subsection{Sparse Bayesian learning}

In the regression setting that we are considering next, data consists of a set of input and output points
\begin{equation}
\calD := \left\{ \left(\bxi^{(n)}, y^{(n)} := f(\bxi^{(n)})\right)\right\}_{n=1}^N, \ \ \bxi^{(n)} \sim p_{\bxi}(\cdot)
\end{equation}
and the model that we attempt to fit on the data can be generally written as a linearly-weighted sum of basis functions $\boldsymbol{\Phi}(\bxi) = (\phi_1(\bxi), \dots, \phi_M(\bxi))^T$, with weights $\bw = (\ww_1, \dots, \ww_M)^T$, that is
\begin{equation}
y = \sum_{i=1}^{N_K} \ww_i \phi_i (\bxi) = \bw^T \boldsymbol{\Phi}(\bxi).
\end{equation}
The above setting is reminiscent of the support vector machine models \cite{boser, vapnik}, although here we do not impose any restrictions on the basis functions to be kernel functions that satisfy Mercer's condition. Instead, our goal is to explore a fully Bayesian approach for determining ``good" estimates for $\bw$ that include relatively few non-zero values indicating the basis functions that are the most ``relevant" for making good predictions. First, by taking liberty on the choice of basis functions, we utilize the PCE expansion introduced above, that is we choose the $\phi_i$'s to be orthogonal polynomials. Next, in order to enforce sparsity, we introduce an auxilliary variable $\biota = (\iota_1, \dots, \iota_{M})^T$, that is a vector of binary components that admit values $0$ or $1$, thus, determining whether to discard the corresponding basis term or not. This way the model is weighted using $\ww_i\iota_i$ and for each input point, we rewrite the output representation as an expansion of the form 
\begin{equation}
\label{eq:pce_sparse}
y^{(n)} := \sum_{i = 1}^{N_{K}} \ww_i \iota_i \Psi_{i}(\bxi^{(n)}) = \left(\bw \circ \biota\right)^T \bPsi(\bxi^{(n)}) , \ \ n = 1, \dots, N,
\end{equation}
where $\bPsi(\bxi^{(n)}) = \left[\Psi_1(\bxi^{(n)}), \dots ,\Psi_{N_K}(\bxi^{(n)})\right]^T$. In the above, we denote with ``$\circ$" the component-wise (Hadamard) product and we refer to $\bw \circ \biota$ as the \emph{relevance vector}. Sparsity is in fact introduced in the above model by appropriately restricting the number of $\biota$ entries that are nonzero to be much smaller than $N_K$, in a way that we describe below.

In order to infer the unknown chaos coefficients $\bw$ and the values of the associated $\biota$ in (\ref{eq:pce_sparse}), we start by assuming a Gaussian noise model for describing the observations error.
That is, given the coefficients $\bw$, and the sparsity indices $\biota$, each observation $y^{(n)}$ is normally distributed with probability density given as 
\begin{equation}
p(y^{(n)} \vert \bw, \biota, \bxi^{(n)}, \tau) = \calN(y^{(n)} \vert \left(\bw \circ \biota\right)^T \bPsi(\bxi^{(n)}), \tau^{-1}),
\end{equation}
$\calN(\cdot \vert \mu, \sigma^2)$ is the Gaussian probability density with mean $\mu$ and variance $\sigma^2$, and $\tau$ is the inverse variance or precision parameter. Assuming that the data are generated by independent sampling, the joint conditional distribution of $\{y_n\}_{n=1}^N$ becomes
\begin{equation} \label{eq:likelihood}
p(\by \vert \bw, \biota, \bXi, \tau) = \prod_{n=1}^N \calN(y^{(n)} \vert \left(\bw \circ \biota\right)^T \bPsi(\bxi^{(n)}), \tau^{-1}) = \frac{1}{(2\pi\tau^{-1})^{N/2}}\exp\left\{ -\frac{1}{2}\tau \vert\vert \by - \bPsi(\bXi) (\bw\circ \biota) \vert\vert^2\right\},
\end{equation}
where, for convenience, we write $\bXi = \{\bxi^{(n)}\}_{n=1}^N$ and $\bPsi(\bXi)$ is the $N\times N_{K}$ matrix with entries $\Psi_j(\bxi^{(i)})$, $i = 1, \dots, N$, $j = 1, \dots, N_K$. 
Next, we assign independent, zero mean Gaussian prior distributions to the components of $\bw$, that is  
\begin{equation}
p(\bw \vert \bvsigma) := \prod_{i=1}^{N_K} \calN\left( \ww_i \vert 0, \varsigma_i^{-1}\right) = \prod_{i=1}^{N_K} \frac{1}{(2\pi \varsigma_i^{-1})^{1/2}} \exp\left\{ - \frac{\varsigma_i \ww_i^2}{2} \right\}
\end{equation}
where the inverse variance parameters $\{\varsigma_i\}_{i=1}^{N_K}$ are modelled as iid random variables that follow a Gamma prior distribution with parameters $a$ and $b$,
\begin{equation}
p(\bvsigma \vert a, b) := \prod_{i=1}^{N_K} \calG(\varsigma_i \vert a,b ) = \left( \frac{b^a}{\Gamma(a)} \right)^{N_K} \prod_{i=1}^{N_K} \varsigma_i^{a-1}e^{-b\varsigma_i}.
\end{equation}
For the above choices of priors for $\bw$ and $\bvsigma$, one can obtain the overall (predictive) prior for $\bw$ by marginalizing over $\bvsigma$ as 
\begin{equation}
p(\bw \vert a, b) = \prod_{i=1}^{N_K} \int \calN(\ww_i \vert 0, \varsigma_{i}^{-1})\calG(\varsigma_i \vert a, b) d \varsigma_i,
\end{equation}
which results in an independent Student-t distribution and can be used to promote sparsity in the solution when $a$, $b$ are assigned to very small values. Specifically, the case $a = b = 0$ corresponds to uninformative priors $p(\ww_i) \propto 1/\vert\ww_i\vert$ that concentrate sharply around zero \cite{tipping}, similar to the Laplace priors used in \cite{sargsyan_IJUQ}.

To the sparsity indices $\biota$ we assign independent Bernoulli priors with success probabilities $\bpi:= (\pi_1, \dots, \pi_{N_K})^T$
\begin{equation}
p(\biota \vert \bpi) := \prod_{i=1}^{N_K} \sfB(\iota_i \vert \pi_i) = \prod_{i=1}^{N_K} \pi_i^{\iota_i} (1-\pi_i)^{1 - \iota_i}
\end{equation}
where $\bpi$ follows a product of Beta prior distributions
\begin{equation}
p(\bpi \vert c,d) := \prod_{i=1}^{N_K} \calB(\pi_i \vert c,d) = \left(\frac{1}{B(c, d)}\right)^{N_K} \prod_{i=1}^{N_K} \pi_i^{c-1} (1 - \pi_i)^{d - 1}.
\end{equation}
In this case, by marginalizing over the hyperparameters $\bpi$ one obtains the prior 
\begin{equation}
p(\biota \vert c, d) = \prod_{i=1}^{N_K}\int_0^1 p(\iota_i \vert \pi_i)p(\pi_i \vert c, d) d\pi_i = \prod_{i=1}^{N_K}\left(\frac{c}{c+d}\right)^{\iota_i}\left(\frac{d}{c+d}\right)^{1 - \iota_i},
\end{equation}
that is a product of Bernoulli densities with success probability $c/(c+d)$. As will be demonstrated in the numerical examples later, the choice of $c$, $d$ parameters allows for controling the level of sparsity in the model. Specifically, by taking $c$ to be close to zero, the probability that the $\iota_i$ will be one and therefore for the corresponding term to be relevant in the model becomes very small. As a result of this, the most important components will have a posteriori high success probabilities, while the rest will remain close to zero. At last, the noise precision parameter $\tau$ that appears in $p(\by \vert \bc, \biota, \bXi, \tau)$ is assigned a Gamma distribution 
\begin{equation}
p(\tau \vert u, w) = \calG(\tau \vert u,w).
\end{equation}
Putting everything together, one can write the posterior distribution of all parameters $\btheta = \{\bw, \bvsigma, \biota, \bpi,\tau\}$ conditioned on $\calD$ as 
\begin{equation}
\label{eq:post}
p_{\btheta \vert \calD}(\btheta) \propto p_{\calD, \btheta}(\by,\btheta) := p(\by \vert \bw, \biota, \bXi, \tau) p(\bw \vert \bvsigma) p(\bvsigma \vert a, b) p(\biota \vert \bpi) p(\bpi \vert c,d) p(\tau \vert u, w).
\end{equation}
A schematic of the Bayesian framework is illustrated as a graphical model in Fig. \ref{fig:graph_model}. Characterizing the above posterior density is not a trivial task. Previous attempts in the RVM and Bayesian Compressive Sensing literature include optimizing the model hyperparameters by maximizing the marginal likelihood \cite{tipping, bilionis_RVM, sargsyan_IJUQ}. As opposed to obtaining point estimates using this approach, in our case we seek a more complete Bayesian treatment as in \cite{bishop_tipping}. It is important to note that the total number of parameters is $N_{\btheta} = 4N_{K} + 1$ where $N_K$ increases factorially as a function of $K$ and $P$.
Therefore, employing sampling techniques such as standard Markov Chain Monte Carlo samplers is extremely inefficient. In the next section, we describe the variational inference framework that will be used in order to approximate (\ref{eq:post}).

\begin{figure}[h]
	\centering
    \tikz{ %
    \node[const] (ab) {$a, b$};
    \node[infer, right=of ab] (omega) {$\boldsymbol{\varsigma}$};
    \node[infer, right=of omega] (coeff) {$\bw$}; 
    \node[infer, above=of coeff, yshift = -.5cm] (tau) {$\tau$};
    \node[const, left=of tau] (uw) {$u, w$};
    \node[infer, right=of coeff] (chaos) {$p(\by \vert \bw, \biota, \bXi, \tau) $};
    \node[infer, below=of coeff, yshift = 0.5cm] (sparse) {$\biota$};
    \node[infer, left=of sparse] (pi) {$\bpi$};
    \node[const, left=of pi] (cd) {$c, d$};
    \node[output, below=of chaos] (data) {$\mathcal{D} = \{\bXi, \by\}$};
    \edge {ab} {omega} ;
    \edge {omega} {coeff};
    \edge {uw} {tau} ;
    \edge {coeff} {chaos};
    \edge {tau} {chaos};
    \edge {data} {chaos};
    \edge {sparse} {chaos};
    \edge {pi} {sparse};
    \edge {cd} {pi};
    }
    \caption{Graphical model representation of the inference framework. The likelihood function is characterized by the data $\calD$, the chaos coefficients $\bw$, the sparsity indices $\biota$ and the noise scale $\tau$. The $\bw$ and $\biota$ are characterized by their affiliated parameters $\bvsigma$ and $\bpi$ respectively, whose priors are specified by the constants $(a,b)$ and $(c,d)$, while $\tau$ is specified using $(u,w)$.\label{fig:graph_model}}
\end{figure}

\section{Variational inference}
\label{sec:VI}

\subsection{Optimization goals}

We approximate the posterior distribution given in eq. (\ref{eq:post}) using a
variational approach. Consider a family of parametric densities $\left\{q_{\btheta|\bell}(\btheta | \bell)\right\}_{\bell \in \boldsymbol{\Lambda}}$, where $\boldsymbol{\Lambda}$ is the set of admissible parameter vectors. We aim at determining the parameters $\bell \in \boldsymbol{\Lambda}$ such that $q_{\btheta|\bell}(\btheta | \bell)$ is sufficiently ``close" to our target posterior. This proximity is quantified by the Kullback-Liebler divergence \cite{kullback}.  Specifically, we want to solve the minimization problem 
\begin{equation}
\bell^* = \underset{\bell \in \boldsymbol{\Lambda}}{\arg\min}~ \KL\left[q_{\btheta|\bell} \big\vert \big\vert p_{\btheta|\calD} \right], 
\end{equation}
where 
\begin{equation}
\KL\left[q_{\btheta|\bell} \big\vert\big\vert p_{\btheta |
    \calD}\right] = \int q_{\btheta | \bell}
(\btheta | \bell) \log \left[ \frac{q_{\btheta|\bell} (\btheta \vert\bell) }{p_{\btheta | \calD} (\btheta)} \right] d\btheta.
\end{equation}
Using Bayes' rule to expand the posterior $p_{\btheta | \calD} (\btheta)$ as in eq. (\ref{eq:post}) and denoting with $p_{\calD}(\by)$ the normalizing constant, one gets the identity
\begin{equation}
\label{eq:evidence_ident}
\log p_{\calD}(\by) = \calF[q_{\btheta|\bell}] + \KL\left[ q_{\btheta|\bell} \big\vert \big\vert p_{\btheta|\calD} \right],
\end{equation}
with 
\begin{equation}
\calF[q_{\btheta\vert \bell}] = \E_{q_{\btheta\vert\bell}}[\log p_{\calD, \btheta}(\by, \btheta)] + \calH[q_{\btheta\vert \bell}],
\end{equation}
and
\begin{equation}
\calH[q_{\btheta\vert\bell}] = -\int q_{\btheta\vert\bell}(\btheta\vert\bell) \log q_{\btheta\vert\bell}(\btheta\vert \bell) d\btheta,
\end{equation}
where $p_{\calD, \btheta}(\by, \btheta)$ is the right hand side expression in eq. (\ref{eq:post}).
From (\ref{eq:evidence_ident}) we can see that the left hand side is
the log-evidence quantity that is fixed for a certain dataset $\calD$, therefore minimizing the KL divergence is
equivalent to maximizing $\calF[q_{\btheta\vert\bell}]$. The latter is called
\emph{evidence lower bound} and is the sum of the expectation of the
log-joint density $\log p_{\calD, \btheta}(\by, \btheta)$ with respect to $q_{\btheta\vert\bell}$ and the entropy $\calH[q_{\btheta \vert \bell}]$ of the approximating distribution. 

\subsection{Posterior factorization within the prior exponential families}

We remind that the target distribution of $\btheta$ is $p_{\btheta | \calD}(\btheta) \propto p_{\calD,\btheta}(\by, \btheta) = p_{\calD|\btheta}(\by\vert\btheta) p_{\btheta}(\btheta)$,
where from eq. (\ref{eq:post}) it follows that the likelihood term is $p_{\calD\vert\btheta}(\by
\vert\btheta):= p(\by \vert \bw, \biota, \bXi, \tau)$ and the prior of $\btheta$ is 
\begin{eqnarray}
\label{eq:prior_fact}
p_{\btheta}(\btheta) & = & p(\bw \vert \bvsigma) \ 
                            p(\bvsigma \vert a, b) \ 
                            p(\biota \vert \bpi) \
                            p(\bpi \vert c,d) \ 
                            p(\tau \vert u, w)
                           \nonumber \\ & = & 
\left( \prod_{i=1}^{N_{K}} \calN(\ww_i | 0, \varsigma_i^{-1}) \right) \left( \prod_{i=1}^{N_{K}} \calG(\varsigma_i | a, b) \right) \left( \prod_{i=1}^{N_{K}} \sfB( \iota_i | \pi_i)\right) \left( \prod_{i=1}^{N_{K}} \calB(\pi_i | c, d) \right) \calG(\tau | u, w).
\end{eqnarray} 
The prior choices for $\btheta$, shown above, are exponential family distributions whose probability densities can all be written in their \emph{canonical} form 
\begin{equation}
p_{\theta_i}(\theta_i) = h(\theta_i) \exp\left\{\bzeta_i^T R(\theta_i) - A_i(\bzeta_i) \right\},
\end{equation}
where $\bzeta_i$ is the natural parameter, $R(\theta_i)$ is a vector valued function of $\theta_i$ that constitutes a sufficient statistic and $A_i(\bzeta_i)$ is the log of the normalizing factor. Their explicit expressions for each distribution are given in \ref{sec:appA}.

We define the approximating family of distributions to consist of probability densities $q_{\btheta|\bell}(\btheta)$, parameterized by a parameter vector $\bell$, that can also be factorized as in (\ref{eq:prior_fact}), that is 
\begin{eqnarray}
\label{eq:post_fact}
q_{\btheta | \bell}(\btheta ) &=& q_{\bw}(\bw) \ 
                                q_{\bvsigma}(\bvsigma) \ 
                                q_{\biota}(\biota) \ 
                                q_{\bpi}(\bpi) \ 
                                q_{\tau}(\tau)
                                \nonumber \\ & = &
\left( \prod_{i=1}^{N_{K}} q_{\ww_i}(\ww_i) \right) \left(\prod_{i=1}^{N_{K}} q_{\varsigma_i}(\varsigma_i) \right) \left(\prod_{i=1}^{N_{K}} q_{\iota_i}(\iota_i) \right) \left(\prod_{i=1}^{N_{K}} q_{\pi_i}(\pi_i) \right) q_{\tau|r,s}(\tau),
\end{eqnarray}
where
\begin{equation}
q_{\ww_i|m_i,\rho_i}(\ww_i) = \calN(\ww_i | m_i, \rho_i),
\end{equation}
\begin{equation}
q_{\varsigma_i|\kappa_i,\lambda_i}(\varsigma_i) = \calG(\varsigma_i | \kappa_i, \lambda_i),
\end{equation}
\begin{equation}
q_{\iota_i | \tilde{\pi}_i}(\iota_i) = \sfB(\iota_i | \tilde{\pi}_i),
\end{equation}
\begin{equation}
q_{\pi_i| r_i, s_i} = \calB(\pi_i | r_i, s_i),
\end{equation}
\begin{equation}
q_{\tau|\upsilon,\omega}(\tau) = \calG(\tau | \upsilon, \omega).
\end{equation}
Thus, the distributions of the components $q_{\theta_i}(\theta_i)$ are in the same exponential family as their corresponding priors, that is, their canonical forms
\begin{equation}
\label{eq:exp_fam_post}
q_{\theta_i}(\theta_i) = h(\theta_i) \exp\left\{ \bfeta_i^T R(\theta_i) - A_i(\bfeta_i)\right\},
\end{equation}
are characterized by the same sufficient statistic $R(\cdot)$, log-normalizing constant $A_i(\cdot)$ and function $h(\cdot)$ and only the natural parameter $\bfeta_i$ differs.

The above factorized density (\ref{eq:post_fact}) is parameterized by $\bell = \{m_i, \rho_i\}_{i=1}^{N_{K}} \cup \{\kappa_i, \lambda_i\}_{i=1}^{N_{K}} \cup \{\tilde{\pi}_i\}_{i=1}^{N_{K}} \cup \{r_i, s_i\}_{i=1}^{N_{K}} \cup \{\upsilon, \omega \}$ and our objective is to find $\bell^*$ that maximizes $\calF[q_{\btheta|\bell}]$. The main advantage of restricting ourselves to a factorized posterior density in the same exponential family as the prior, is that it allows factorization of the integrals involved in $\calF[q_{\btheta|\blambda}]$ and from that its analytical computation. Furthermore, the resulting optimization problem in this approach, known also as \emph{mean field variational inference} \cite{jordan}, becomes a convex problem and can be approached using a batch parameter updating scheme \cite{winn}, as will be seen below. In addition to being crucial for enabling the analytical computation of the objective function, independence among the coefficients simplifies their posterior marginal density representation, which suffices for point predictions and tight confidence intervals on the parameter values. At this point we are not particularly interested in exploring possible interdependencies among the chaos coefficients. The specific expressions for $R(\theta)$ and $A(\bfeta)$ as well as the relation between the parameters in the canonical and non-canonical forms for the Gaussian, Gamma, Bernoulli and Beta distributions are provided in \ref{sec:appA}.

We remark that the variational approach presented herein does not take advantage of the analytical solution to the posterior distribution of the coefficients $\bw$ conditional on the parameters $\{\bvsigma, \biota, \tau\}$.
This distribution is readily available because the chosen prior of $\bw$ is a conjugate prior for the normal likelihood of eq. \eqref{eq:likelihood}, e.g. see \cite{tipping}. 
A variational approach could be applied to approximate the distribution of  $\{\bvsigma, \biota, \bpi,\tau\}$ conditional on $\calD$ using the marginal likelihood obtained through integrating out $\bw$ from eq. \eqref{eq:post}.
However, such an approach would require an additional step to obtain the marginal posterior of  $\bw$ through marginalising out $\{\bvsigma, \biota, \bpi,\tau\}$ from the joint posterior.
In contrast, the presented approach gives directly the marginal posterior of $\bw$ due to the factorized form of the parametric density of \eqref{eq:post_fact}.

\subsubsection{Computation of Entropy}

First we need to compute 
\begin{eqnarray}
\calH[q_{\btheta\vert \bell}]  & = & - \int q_{\btheta\vert\bell}(\btheta) \log q_{\btheta\vert\bell}(\btheta) d\btheta  \nonumber \\ & = & 
- \int  \prod_{i=1}^{N_{\btheta}}q_{\theta_i}(\theta_i) \left[\sum_{i=1}^{N_{\btheta}} \log q_{\theta_i}(\theta_i)\right] \prod_{i=1}^{N_{\btheta}} d\theta_i \nonumber \\ & = & 
- \sum_{i=1}^{N_{\btheta}} \int q_{\theta_i}(\theta_i) \log q_{\theta_i}(\theta_i) d\theta_i \nonumber\\ & = &
\sum_{i=1}^{N_{\btheta}} \calH[q_{\theta_i}].
\end{eqnarray}
Since all $q_{\theta_i}$ are of the same exponential family distribution as in the prior case, given in eq. (\ref{eq:exp_fam_post}), and $\bfeta := \bfeta(\bell)$ is 
the natural parameter, each individual entropy term can be written as 
\begin{equation}
\calH[q_{\theta_i}] =  -\E_{q_{\theta_i}}[\log h(\theta_i)] + A_i(\bfeta) - \bfeta^T \nabla_{\bfeta} A_i(\bfeta), \ \ i = 1, \dots, N_{\btheta}
\end{equation}
where we used the property of sufficients statistics $\E_{q_{\theta_i}}[R(\theta_i)] = \nabla_{\bfeta} A_i(\bfeta)$.

\subsubsection{Computation of the expected log joint distribution}

Writing $\E_{q_{\btheta\vert\bell}}[\log p_{\calD,\btheta}(\by, \btheta)] = \E_{q_{\btheta\vert\bell}}[\log
p_{\calD | \btheta}(\by\vert\btheta)] + \E_{q_{\btheta\vert\bell}}[\log p_{\btheta}(\btheta)]$, we first have 
\begin{eqnarray}
\!\!\!\!\!\!\!\! \!\!\!\!\!\!\!\! \E_{q_{\btheta\vert\bell}}[\log p_{\btheta}(\btheta)] & = & \int
                                                     q_{\btheta\vert\bell}(\btheta)
                                                     \log
                                                     p_{\btheta}(\btheta)
                                                     d\btheta \nonumber \\
  & = & \int q_{\bw}(\bw) q_{\bvsigma}(\bvsigma) q_{\biota}(\biota) q_{\bpi}(\bpi) q_{\tau}(\tau)
                  \log \left[p(\bw\vert\bvsigma) p(\bvsigma) p(\biota\vert\bpi) p(\bpi)
                  p(\tau)\right] d\bw d\bvsigma d\biota d\bpi d\tau \nonumber \\ & = & 
\int q_{\bw}(\bw)q_{\bvsigma}(\bvsigma)\log p(\bw \vert \bvsigma)d\bw d\bvsigma +\int q_{\bvsigma}(\bvsigma) \log p(\bvsigma) d\bvsigma + \\ & + & \int q_{\biota}(\biota)q_{\bpi}(\bpi)\log p(\biota\vert\bpi)d\biota d\bpi + \int q_{\bpi}(\bpi) \log p(\bpi)d\bpi + \int q_{\tau}(\tau) \log p(\tau)d\tau.
\end{eqnarray}
In the above we get the general expression
\begin{equation}
\int q_{\theta_i}(\theta_i) \log p(\theta_i) d\theta_i = \E_{q_{\theta_i}}[\log h(\theta_i)] + \bzeta_{i}^T \nabla A_{i}(\bfeta_{i}) - A_{i}(\bzeta_{i})
\end{equation}
for $\theta_i \in \{\varsigma_j\}_{j=1}^{N_{K}} \cup \{\pi_j\}_{j=1}^{N_{K}} \cup \{\tau\}$. For the remaining terms we get
\begin{eqnarray}
\!\!\!\!\!\!\!\! \!\!\!\!\!\!\!\! \int
  q_{\bw}(\bw)q_{\bvsigma}(\bvsigma)\log p(\bw|\bvsigma) d\bw
  d\bvsigma & = & \sum_{i=1}^{N_{K}} \int
                 q_{\ww_i}(\ww_i)q_{\varsigma_i}(\varsigma_i)\log
                 p(\ww_i \vert \varsigma_i) d\ww_i d\varsigma_i \nonumber \\ & = & \sum_{i=1}^{N_{K}}\left( -\log 2\pi + \E_{q_{\varsigma_i}}[\bzeta_{\ww_i}]^T \nabla A_{\ww}(\bfeta_{\ww_i}) - \E_{q_{\varsigma_i}}[A_{\ww}(\bzeta_{\ww_i})] \right)
\end{eqnarray}
and
\begin{eqnarray}
\!\!\!\!\!\!\!\! \!\!\!\!\!\!\!\! \int
  q_{\biota}(\biota)q_{\bpi}(\bpi)\log p_{\biota|\bpi}(\biota) d\biota
  d\bpi & = & \sum_{i=1}^{N_{K}} \int
                 q_{\iota_i}(\iota_i)q_{\pi_i}(\pi_i)\log
                 p(\iota_i\vert\pi_i) d \iota_i d\pi_i \nonumber \\ & = & \sum_{i=1}^{N_{K}}\left(  \E_{q_{\pi_i}}[\bzeta_{\iota_i}]^T \nabla A_{\iota}(\bfeta_{\iota_i}) - \E_{q_{\pi_i}}[A_{\iota}(\bzeta_{\iota_i})] \right).
\end{eqnarray}
Note that $\bzeta_{\theta_i}$ depends on $\varsigma_i$ or $\pi_i$ for $\theta_i \in \{\ww_j\}_{j=1}^{N_{K}} \cup \{\iota_j\}_{j=1}^{N_{K}}$ and its expectation is taken with respect to the corresponding parametric density. Using the expressions for $\bzeta$ and $A(\cdot)$ given in \ref{sec:appA}, we compute explicitly their expectations $\E_{q_{\varsigma_i}}[\bzeta_{\ww_i}] = (0, \frac{\eta_{\varsigma,1}+1}{2\eta_{\varsigma,2}})$ and $\E_{q_{\varsigma_i}}[A_{\ww}(\bzeta_{\ww_i})] = -\frac{1}{2}\phi'(\eta_{\varsigma,1}+1) + \frac{1}{2}\log(-\eta_{\varsigma,2})$ for the Gaussian density $\calN(\ww_i|0, \varsigma_i^{-1})$ and $\E_{q_{\pi_i}}[\zeta_{\iota_i}] = \phi^{(0)}(\eta_{\pi,1}) - \phi^{(0)}(\eta_{\pi,2})$ and $\E_{q_{\pi_i}}[A_{\iota}(\zeta_{\iota_i})] = \phi^{(0)}(\eta_{\pi,1} + \eta_{\pi, 2}) - \phi^{(0)}(\eta_{\pi, 2})$ for the Bernoulli density $\sfB(\iota_i | \pi_i)$, where $\phi^{(0)}(\cdot) := \frac{\Gamma'(\cdot)}{\Gamma(\cdot)}$ is the digamma function.

Next, we have 
\begin{eqnarray}
\!\!\!\!\!\!\!\! \E_{q_{\btheta\vert \bell}}[\log p_{\calD \vert\btheta}(\by\vert\btheta)] & = & \int q_{\bw}(\bw) q_{\biota}(\biota) q_{\tau}(\tau)\left\{ -\frac{N}{2}\log(2\pi\tau^{-1}) - \frac{1}{2}\tau \big\vert \big\vert \by - \bPsi(\bXi)(\bw \circ \biota)\big\vert\big\vert^2\right\} d\tau d\bw d\biota \nonumber\\ & = & 
-\frac{N}{2}\log(2\pi) + \int q_{\bw}(\bw) q_{\biota}(\biota) q_{\tau}(\tau) R(\tau)^T \bL\ d\tau d\bw d\biota \nonumber\\ & = & -\frac{N}{2}\log(2\pi) + \E_{q_{\tau}}[R(\tau)]^T \int q_{\bw}(\bw) q_{\biota}(\biota) \bL \ d\bw d\biota
\end{eqnarray}
where $R(\tau) = (\log\tau, \tau)^T$ and 
\begin{eqnarray}
\bL = \left[\begin{array}{c} L_1 \\ L_2 \end{array}\right] = \left[\begin{array}{c}N/2 \\ -\frac{1}{2}\big\vert \big\vert \by - \bPsi(\bXi)(\bw\circ \biota) \big\vert\big\vert^2 \end{array} \right].
\end{eqnarray}
Using again the sufficiency property we get $\E_{\tau}[R(\tau)] = \nabla_\tau A(\bfeta_\tau)$ while the second expectation gives $\E[L_1] = N/2$ and $\E[L_2] = -\frac{1}{2}\big\vert \big\vert \by - \bPsi(\bXi)(\bm\circ \tilde{\bpi}) \big\vert\big\vert^2 -\frac{1}{2}\trace{\bPsi^T\bPsi\left(\diag{\tilde{\bpi} \circ \brho^{-1}} + \diag{(\bm - \bm\circ\tilde{\bpi})\circ(\bm\circ \tilde{\bpi})}\right)}$, where $\bm= [m_1, \dots, m_{N_{K}}]^T$, $\brho = [\rho_1, \dots, \rho_{N_{K}}]^T$, $\tilde{\bpi} = [\tilde{\pi}_1, \dots, \tilde{\pi}_{N_{K}}]^T$ and $m_i = -\frac{\eta_{\ww_i,1}}{2\eta_{\ww_i,2}}$, $\rho_i = -2\eta_{\ww_i, 2}$, $\tilde{\pi}_i = \frac{e^{\eta_{\iota_i}}}{1 + e^{\eta_{\iota_i}}}$.

\subsection{Optimization using a gradient ascent scheme}

Our goal is to maximize $\calF[q_{\btheta\vert\bell}]$ with respect to the parameters $\bell$. Working in the space of the natural parameters $\{\bfeta_i\}$ allows us to follow the approach in \cite{blei_jordan} and propose an algorithm that updates $\bfeta_i$ iteratively to the values for which the gradient $\nabla_{\bfeta_i} \calF[q_{\btheta\vert \bell}] = 0$. For our specific choices of prior and posterior distributions, analytical expressions for the corresponding $\bfeta_i$ are available. Specifically the parameters $\bfeta_\tau$, $\{\bfeta_{\varsigma_i}\}_{i=1}^{N_{K}}$, $\{\bfeta_{\pi_i}\}_{i=1}^{N_{K}}$, $\{\bfeta_{\ww_i}\}_{i=1}^{N_{K}}$, $\{\eta_{\iota_i}\}_{i=1}^{N_{K}}$ are updated according to
\begin{subequations}
\label{updates}
\begin{align}
\label{eq:tau}
\bfeta_\tau & = \bzeta_\tau + \E[\bL],\\
\label{eq:omega}
\bfeta_{\varsigma_i} & = \bzeta_{\varsigma_i} + \frac{1}{2} \left[\begin{array}{c}1 \\ - \frac{\partial}{\partial\eta_{\ww_i,2}}A_{\ww}(\bfeta_{\ww_i})  \end{array}\right], \\
\label{eq:pi}
\bfeta_{\pi_i} & = \bzeta_{\pi_i} + \left[\begin{array}{c} 0 \\ 1 \end{array}\right] - \left[\begin{array}{c} 1 \\ -1 \end{array}\right] \nabla A_{\iota}(\bfeta_{\iota_i}), \\
\label{eq:c}
\bfeta_{\ww_i} & = \E_{q_{\varsigma_i}}[\bzeta_{\ww_i}] + \bv_i \frac{\partial A_\tau(\bfeta_\tau)}{\partial \eta_{\tau,2}}, \\
\label{eq:z}
\eta_{\iota_{i}} & = \E_{\pi_i}[\zeta_{\iota_i}] + \bu_i \frac{\partial A_\tau(\bfeta_\tau)}{\partial \eta_{\tau,2}},
\end{align}
\end{subequations}
where 
\begin{eqnarray}
\bv_i & = \left[ \begin{array}{c}\by^T \bPsi (\bepsilon_i\circ \tilde{\bpi}) - (\bepsilon_i \circ \tilde{\bpi} )^T \bPsi^T\bPsi (\tilde{\bpi} \circ \bm_{-i}) \\ -\frac{1}{2}\trace{\bPsi^T\bPsi\diag{\bepsilon_i\circ \tilde{\bpi}} }\end{array} \right],
\end{eqnarray}
\begin{eqnarray}
\begin{array}{rl}\bu_i & = \by^T \bPsi(\bepsilon_i \circ \bm) - (\bepsilon_i \circ \bm)^T \bPsi^T \bPsi (\tilde{\bpi} \circ \bm) \\ & - \frac{1}{2}\trace{\bPsi^T\bPsi\left( \diag{\bepsilon_i \circ \brho^{-1}} + \diag{ ((\mathbf{1} - 2\tilde{\bpi}) \circ \bm) \circ (\bepsilon_i\circ\bm) } \right)} \end{array}
\end{eqnarray}
and $A_{\ww}(\cdot)$, $A_{\iota}(\cdot)$ $A_\tau(\cdot)$ are the log-normalizing constants for $q_{\ww_i\vert\varsigma_i}(\ww_i)$, $q_{\iota_i\vert\tilde{\pi}_i}(\iota_i)$, $q_{\tau\vert\upsilon, \omega}(\tau)$ respectively, $\bepsilon_i$ is the unit vector with $1$ at the $i$th position and zero elsewhere, while $\bm_{-i}$ is the vector with all its entries being equal to those of $\bm$ except the $i$th entry that is $0$. Note also that from the choice of priors $p(\bw | \bvsigma)$, $p(\biota | \bpi)$ it follows that $\bzeta_{\ww_i}$ depends on $\varsigma_i$; its expectation in Eq. (\ref{eq:c}) is taken with respect to $q_{\varsigma_i\vert\kappa_i, \lambda_i}(\varsigma_i)$, $\zeta_{\iota_i}$ depends on $\pi_i$ and its expectation in Eq. (\ref{eq:z}) is taken with respect to $q_{\pi_i\vert r_i, s_i}(\pi_i)$. Detailed derivation of the above formulas are given in \ref{sec:appB}. This procedure that iterates between updating one parameter at a time while holding all other parameters fixed, results in a coordinate ascent algorithm that is guaranteed to converge to a local maximum. Furthermore, under the condition that the objective function $\calF[q_{\btheta\vert\bell}]$ is strictly convex (which holds in our case \cite{blei_jordan, bertsekas}), this coincides with the unique global maximum. 

As mentioned before, the total number of parameters to be estimated in the RVM structure is $N_{\btheta}$ and grows fast as a function of the PCE order and dimensionality. Therefore, the algorithm can become slow. In order to accelerate the procedure, we further incorporate the following step while iterating over the parameters: Once the iterating procedure over all $\pi_i$'s terminates, and before continuing to the next parameter update, we test for convergence of $\bpi$. When this is achieved, we proceed with iterating over the components of $\bw$, $\bvsigma$ that correspond to the important $\pi_i$ values only (where the level of importance is specified by the user). This reduces significantly the number of terms that need to be updated, particularly when very sparse solutions are possible, and faster convergence of the algorithm is achieved. The updating procedure is summarized in Algorithm \ref{alg:VI}.

\begin{algorithm}[h]
\caption{Iterative algorithm for posterior parameter update\label{alg:VI}}
\SetKwInOut{Initialize}{Initialize}
\SetKwInOut{Require}{Require}
\DontPrintSemicolon
\Require{Data $\calD$, prior distribution parameters $\{a,b\}$, $\{c,d\}$, $\{u,w\}$, convergence tolerance $\delta$, $\tilde{\bpi}$ convergence tolerance $\delta_{\bpi}$, success probability threshold $\epsilon_{\pi}$.}
\Initialize{Compute $\bPsi$ and set $\bfeta_\tau := \bzeta_\tau$, $\bfeta_{\varsigma_{i}} := \bzeta_{\varsigma_{i}}$, $\bfeta_{\pi_{i}} := \bzeta_{\pi_{i}}$, $\bfeta_{\ww_i} := \E_{q_{\varsigma_i}}[\bzeta_{\ww_i}]$, $\eta_{\iota_{i}} := \E_{q_{\pi_i}}[\zeta_{\iota_{i}}]$, $\tilde{\pi}_i = \frac{e^{\eta_{\iota_i}}}{1 + e^{\eta_{\iota_i}}}$ and active coefficient indices $\calI_{active} := \calI_{active}^0 = \{1, \dotsm, N_{K}\}$.}
\Repeat{relative change in $(\bfeta_\tau, \bfeta_{\varsigma_1}, \dots, \bfeta_{\varsigma_{N_{K}}}, \bfeta_{\pi_1}, \dots, \bfeta_{\pi_{N_{K}}}, \bfeta_{\ww_1}, \dots, \bfeta_{\ww_{N_{K}}}, \eta_{\iota_1}, \dots, \eta_{\iota_{N_{K}}})^T$ is less than $\delta$}{
  Update $\bfeta_\tau$ as in (\ref{eq:tau})\\
  \For{ $i \in \calI_{active}$}{
  Update $\bfeta_{\bvsigma_i}$ as in (\ref{eq:omega})\\
  Update $\bfeta_{\bpi_i}$ as in (\ref{eq:pi})\\
  Recompute $\E_{q_{\varsigma_i}}[\bzeta_{\ww_i}]$ and $\E_{q_{\pi_i}}[\zeta_{\iota_i}]$ \\
  Update $\eta_{\iota_i}$ as in (\ref{eq:z}) \\
  Update $\tilde{\pi}_i = \frac{e^{\eta_{\iota_i}}}{1 + e^{\eta_{\iota_i}}}$\\
  Update $\bfeta_{\ww_i}$ as in (\ref{eq:c}) \\
  Update $m_i = -\frac{\eta_{\ww_i,1}}{2\eta_{\ww_i,2}}$\\ 
  }
  \If{relative change in $\tilde{\bpi}$ is less than $\delta_{\bpi}$}{
  Update $\calI_{active} = \{i \in \calI_{active}^0: \tilde{\pi}_i > \epsilon_{\pi}\}$
  }

}
\end{algorithm}

\section{Numerical Examples}
\label{sec:num_ex}

For the numerical examples we set both tolerance critera for convergence in algorithm \ref{alg:VI} to $\delta = \delta_{\bpi} = 10^{-4}$ and the success probability threshold to $\epsilon_{\pi} = 0.01$. We assign broad Gamma priors for $\bvsigma$ and $\tau$ by setting $a = b = u = w = 10^{-6}$. For the Beta prior on $\bpi$, a common choice in the literature is to take $d = (N_K-1)/N_K$ \cite{paisley, chen_beta, lingbo}.
Here, we fix $d = 1$, that is the limiting case when $N_K$ is large. The influence of different values for $c$ is investigated below. For comparison purposes, we also compute the PCE coefficients with a compressive sensing method that relies on the Douglas-Rachford algorithm \cite{DR_algo_1, DR_algo_2}. In order to assess the accuracy of the PCE's that are obtained using both methods and to compare with the true model, we use the empirical relative mean square error defined as 
\begin{equation}
\widehat{MSE} = \sum_{i = 1}^{N_v} \left(f(\bxi^{(i)}) - f_{pc}(\bxi^{(i)}) \right)^2 \bigg/ \sum_{i=1}^{N_v} f^2(\bxi^{(i)})
\end{equation}
evaluated over a set of $N_v$ validation points. For a comparison of the accuracy between different PCEs $f_{pc}$, $g_{pc}$ (obtained using compressive sensing or RVM methods), their $L_2$ distance is computed as 
\begin{equation}
\mathrm{dist}(f_{pc}, g_{pc})_{L_2} = \sum_{i = 1}^{N_K} \left(\ww_i - \ww_i'\right)^2,
\end{equation} 
where $\{\ww_i\}_{i=1}^{N_K}$ and $\{\ww_i'\}_{i=1}^{N_K}$ are the coefficients of the two expansions.

\begin{figure}[h]
\includegraphics[width = \textwidth]{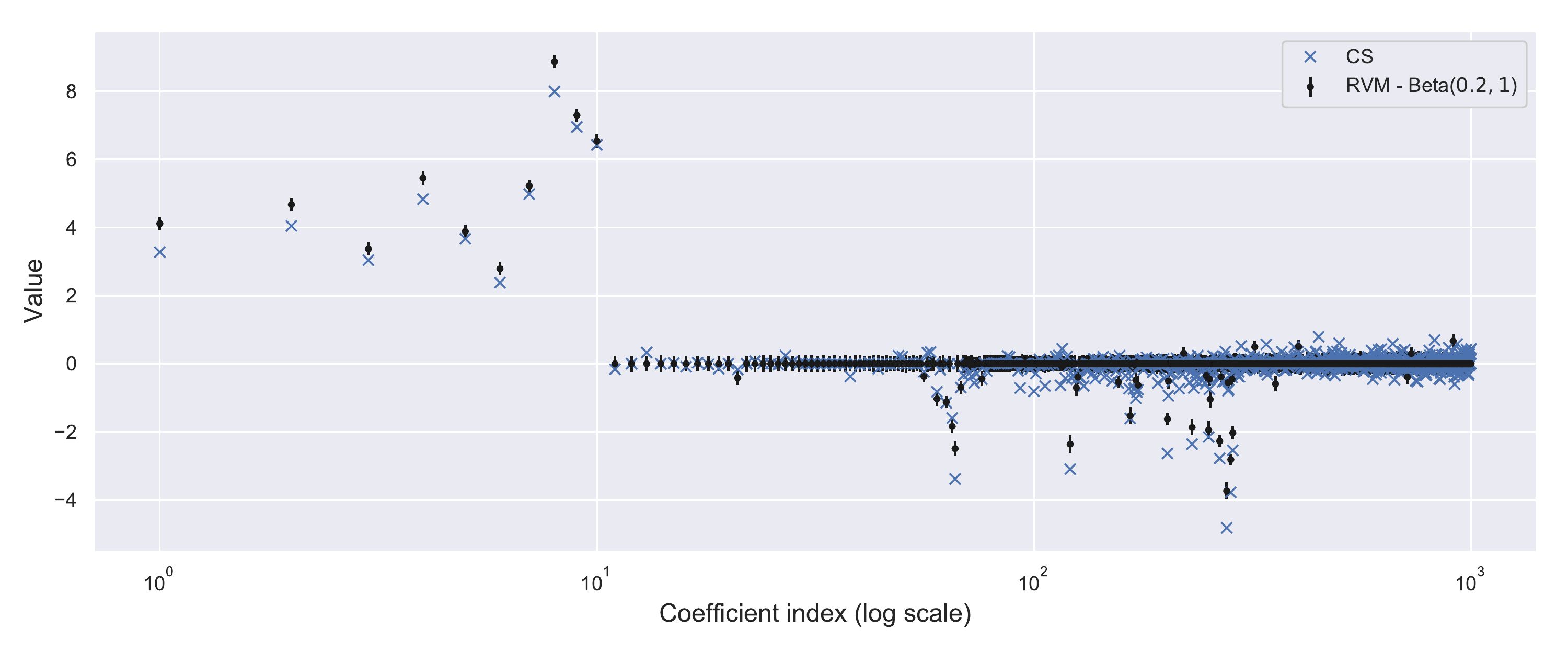}
\includegraphics[width = 0.49\textwidth]{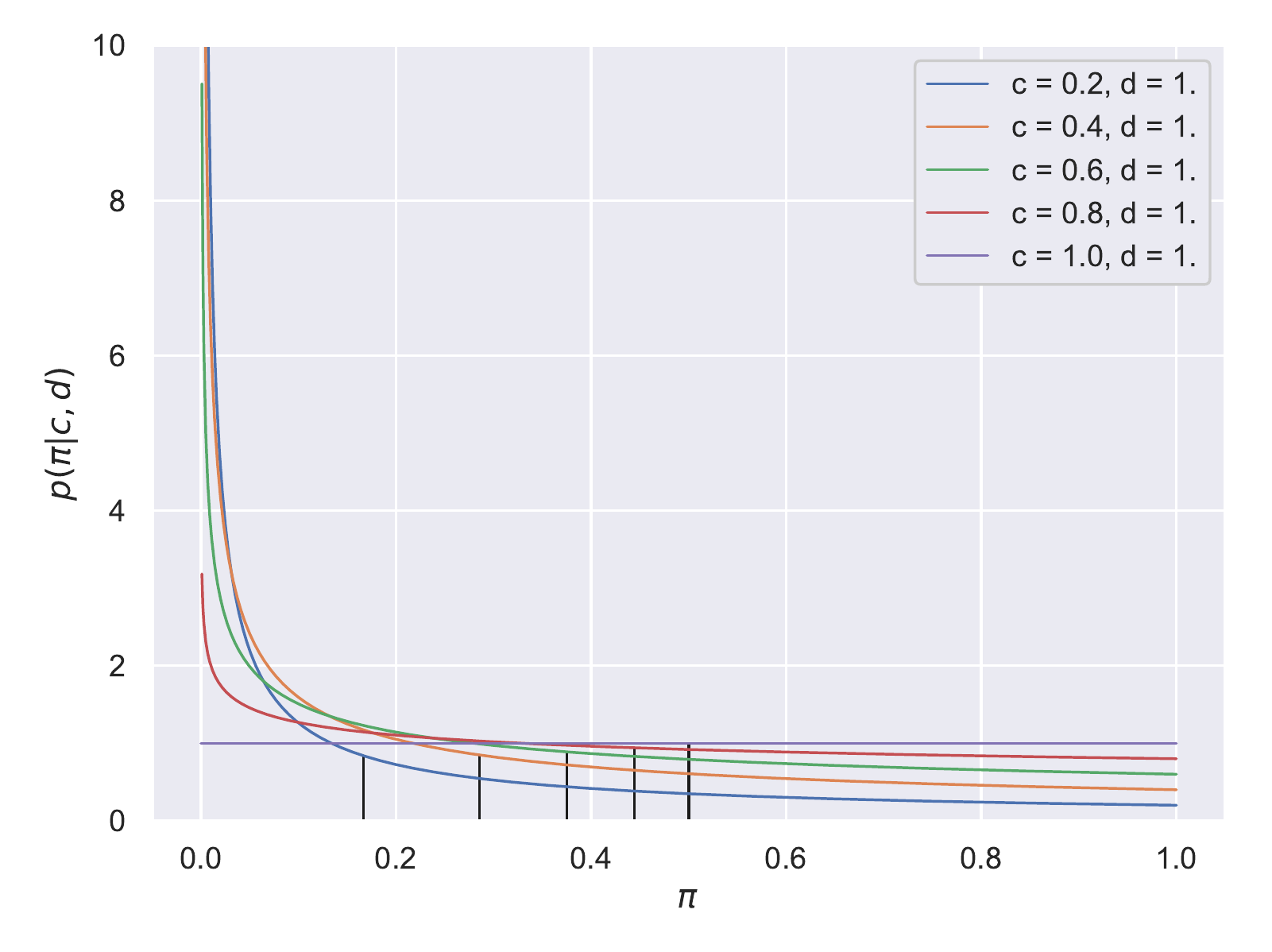}
\includegraphics[width = 0.5\textwidth]{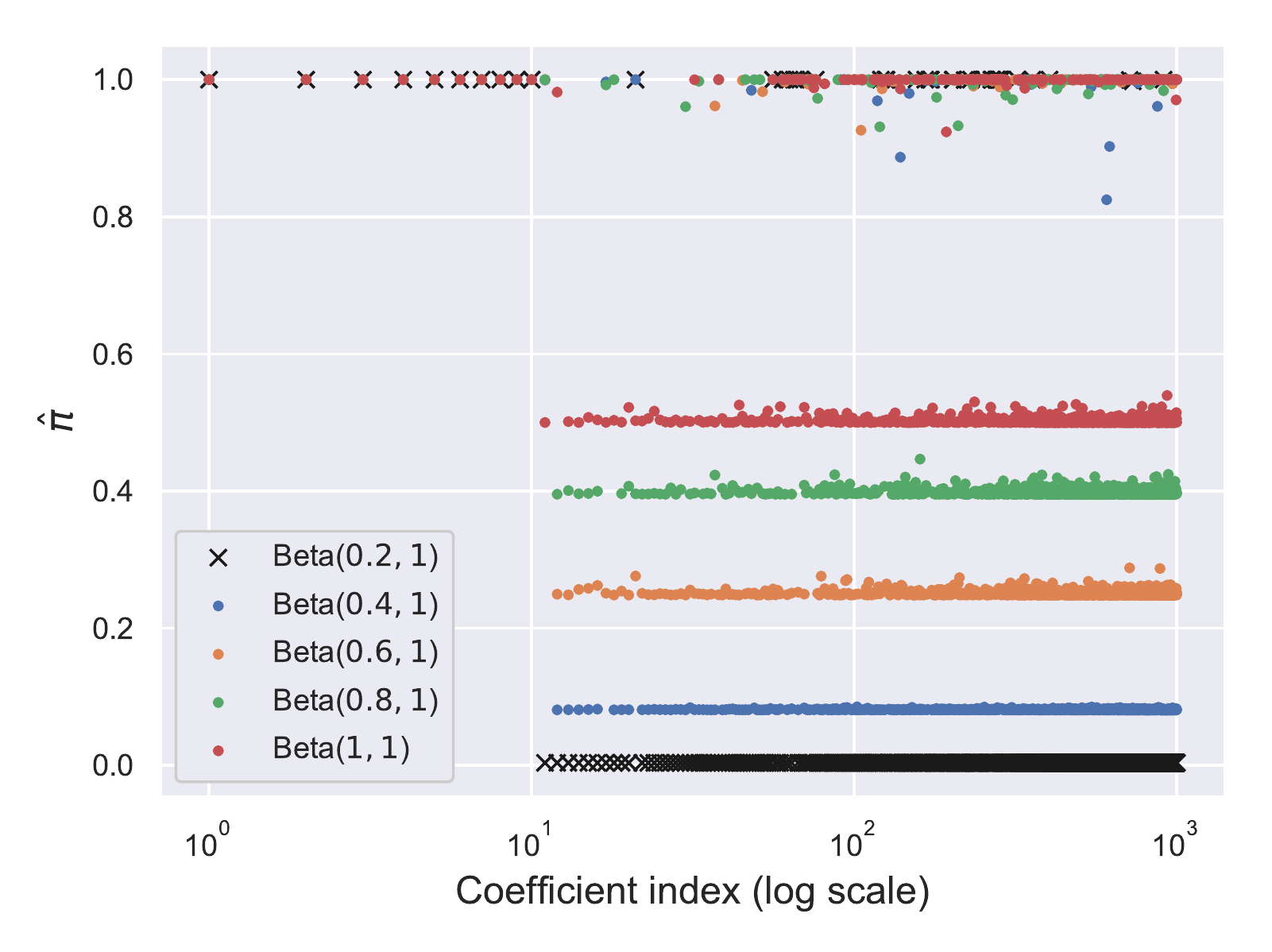}
\caption{Top: Estimates of the coefficients of the PC expansions obtained using compressive sensing (blue 'x' marker) and the proposed relevance vector machine model along with their $2$-standard deviation errorbars (black '.' marker). Bottom left: Plots of the Beta$(\pi|c, d)$ probability density functions for $c = 0.2$, $0.4$, $0.6$, $0.8$, $1$ and $d = 1$. The locations of their means are indicated with black vertical lines. Bottom right: Estimates of the success propability $\hat{\boldsymbol{\pi}}$ obtained using the RVM method for different $c$ parameter of the Beta prior. \label{fig:toy_coeffs}}
\end{figure}

\begin{table}[h]
\caption{Synthetic example - Estimation of the first four statistical moments for PC expansions obtained using varying $c$ parameter in the Beta prior and comparison with true model and PC obtained using CS. The results are compared to MC estimates using the true model with $10^5$ samples. The intervals corresponding to the true model were obtained using bootstrap resampling.}
\label{tab:toy_stats1}
\centering 
\begin{tabular}{l | c | c | c | c | c | c | l}
 & MC & CS & $c=0.2$ & $c = 0.4$ & $c = 0.6$ & $c=0.8$ & $c=1$ \\ 
 \hline 
 \hline 
 Mean & $[6.417, 6.650]$ & $6.515$ & $6.708$ & $6.445$ & $6.644$ & $6.531$ & $6.547$ \\
 Standard deviation & $[18.59, 18.74]$ & $19.79$ & $19.29$ & $19.41$ & $19.17$ & $19.23$ & $19.39$ \\
 Skewness & $[-0.039, -0.015]$ & $-0.013$ & $-0.060$ & $-0.027$ & $-0.032$ & $-0.024$ & $-0.036$ \\
 Kurtosis & $[2.679, 2.718]$ & $2.917$ & $2.791$ & $2.764$ & $2.772$ & $2.738$ & $2.770$ \\
 & & & & & & & \\
 Sparsity Index ($\tilde{\pi}_i > 0.01$) & & & $4.7\%$ & $100\%$ & $100\%$ & $100\%$ & $100\%$ \\
 Sparsity Index ($\tilde{\pi}_i > 0.95$) & & & $4.7\%$ & $10.3\%$ & $12.6\%$ & $15.4\%$ & $15.5\%$ 
\end{tabular}
\end{table}

\subsection{Synthetic example}

We consider the nonlinear model that is described by the modified O'Hagan function \cite{oakley} $f : \R^K \to \R$ given as 
\begin{equation}
f(\bxi) = \ba_1^T\bxi + \ba_2^T\sin(\bxi) + \ba_3^T \cos(\bxi) + \cos(\bxi)^T \bM \sin(\bxi)
\end{equation}
where $\ba_i \in \R^K$, $i =1,\dots, 3$ and $\bM \in \R^{K\times K}$. For this example we fix $K = 10$ and the values of the vector coefficients $\ba_i$ and $\bM$ are randomly generated such that the first $7$ input variables will have a smaller effect than the remaining $3$. Specifically, we generate uniform $\calU(0,1)$ for the first $7$ entries of all $\ba_i$'s and uniform $\calU(1.5, 2)$ for the last $3$. The entries of $\bM$ are all sampled from a uniform $\calU(0,2)$. Reproducibility is enabled by fixing the seed of the random number generator.

\begin{figure}[h]
\centering
\includegraphics[width = 0.49\textwidth]{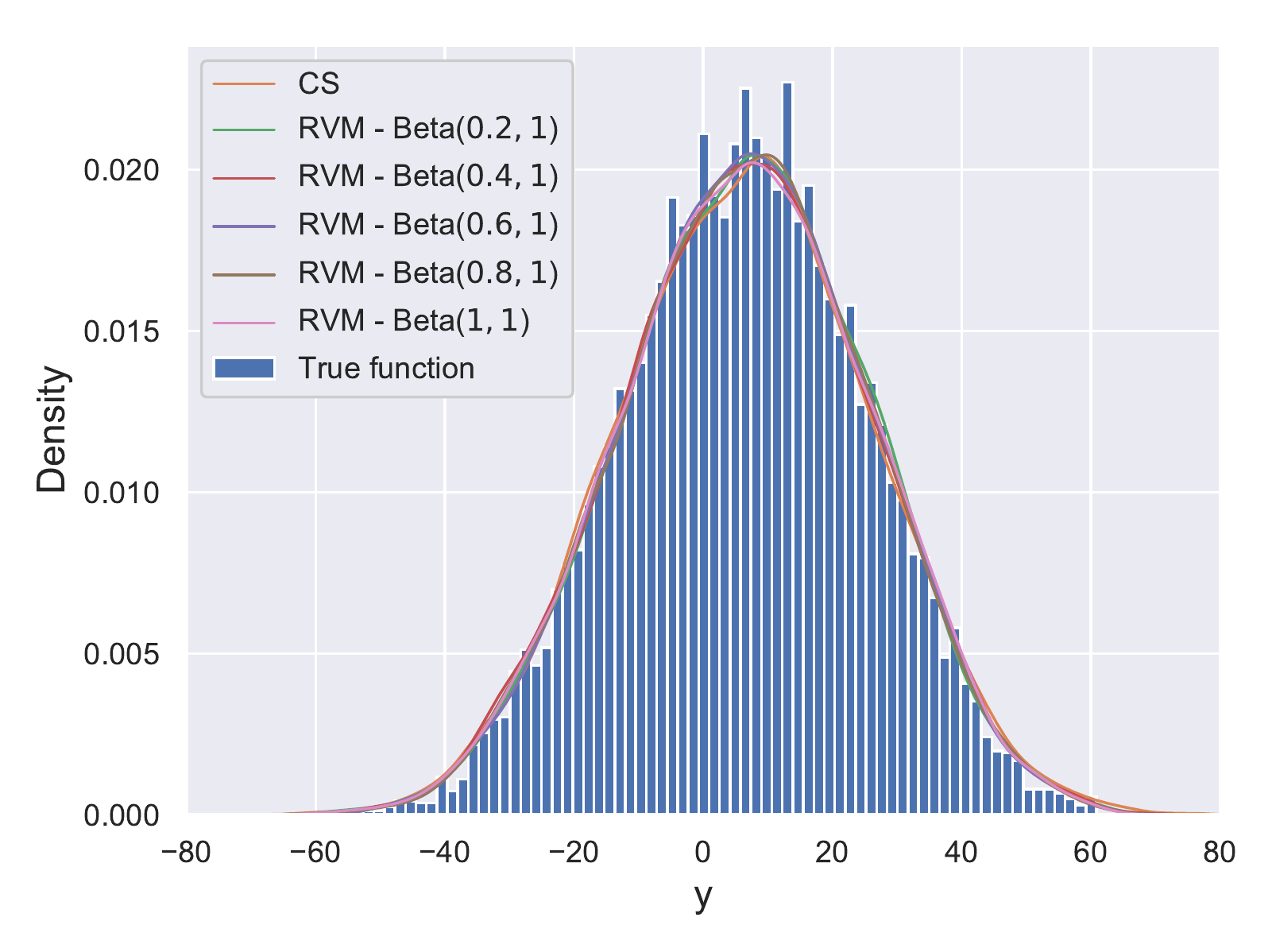}
\includegraphics[width = 0.49\textwidth]{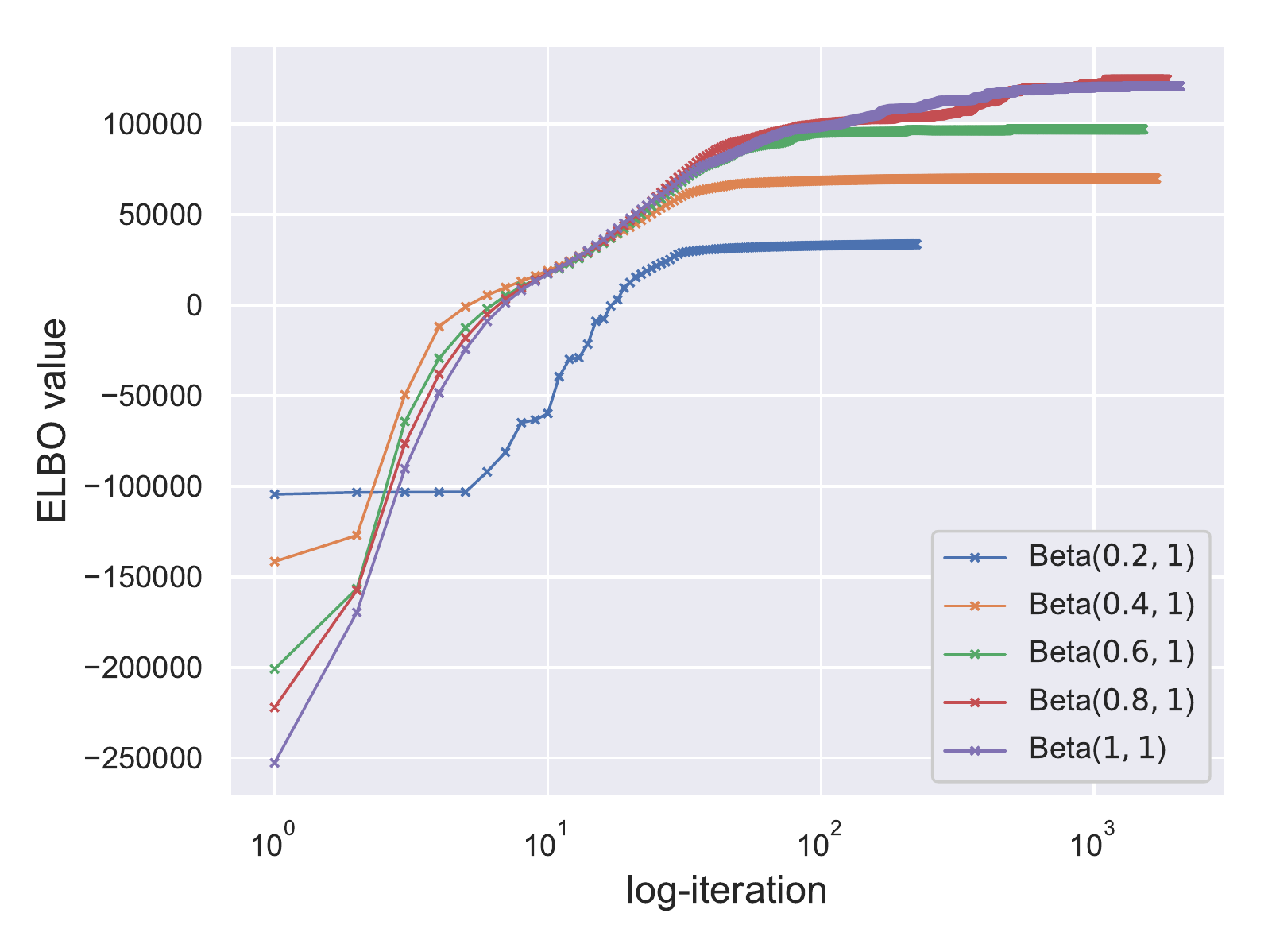}
\caption{Left: Probability densities of the PC expansions obtained using CS and RVM with varying $c$ parameter of the Beta prior. Right: ELBO function evolution vs number of iterations of Algorithm \ref{alg:VI}. \label{fig:pdf_elbo}}
\end{figure}

\begin{figure}[h]
\includegraphics[width = 0.5\textwidth]{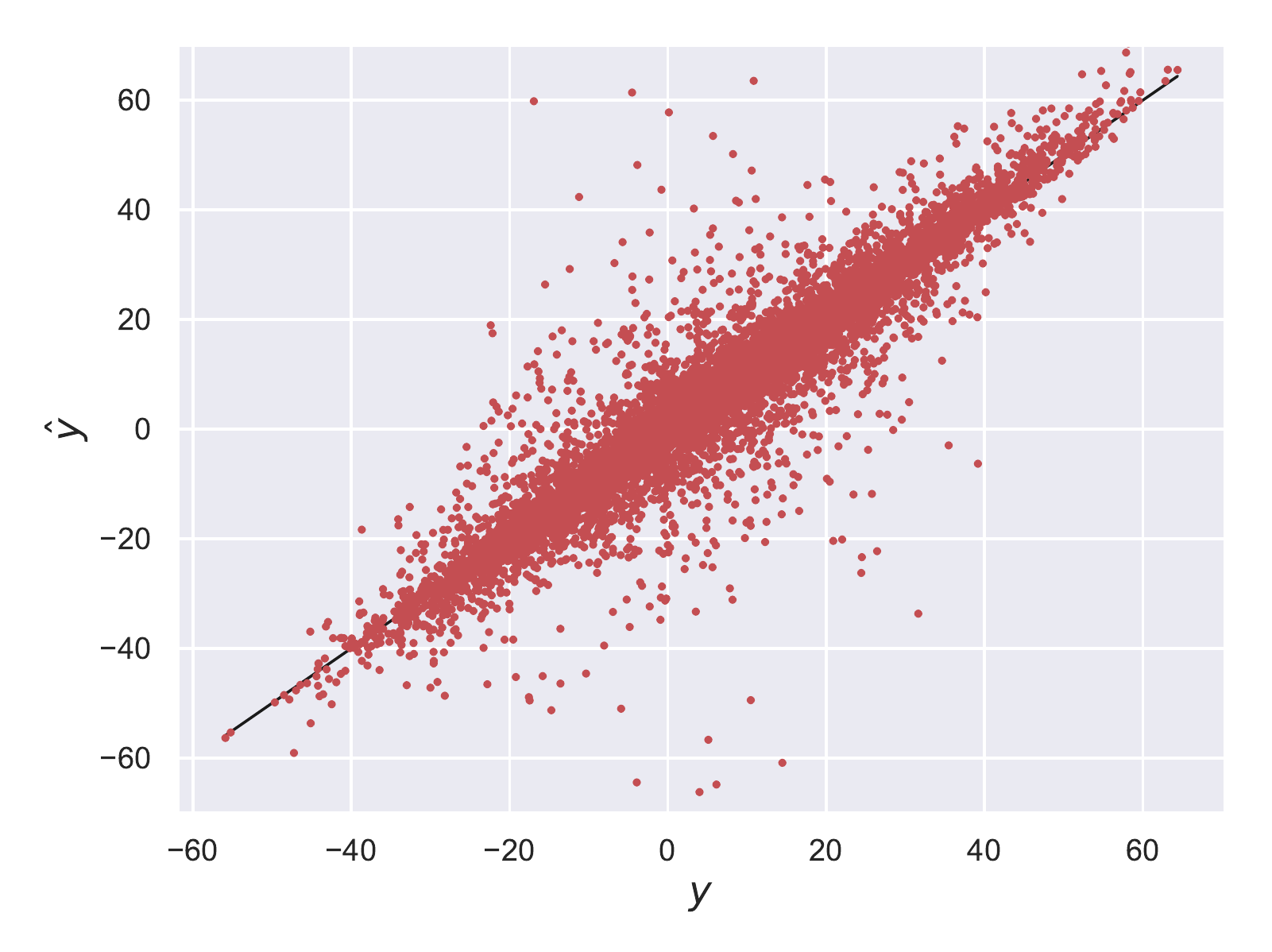}
\includegraphics[width = 0.5\textwidth]{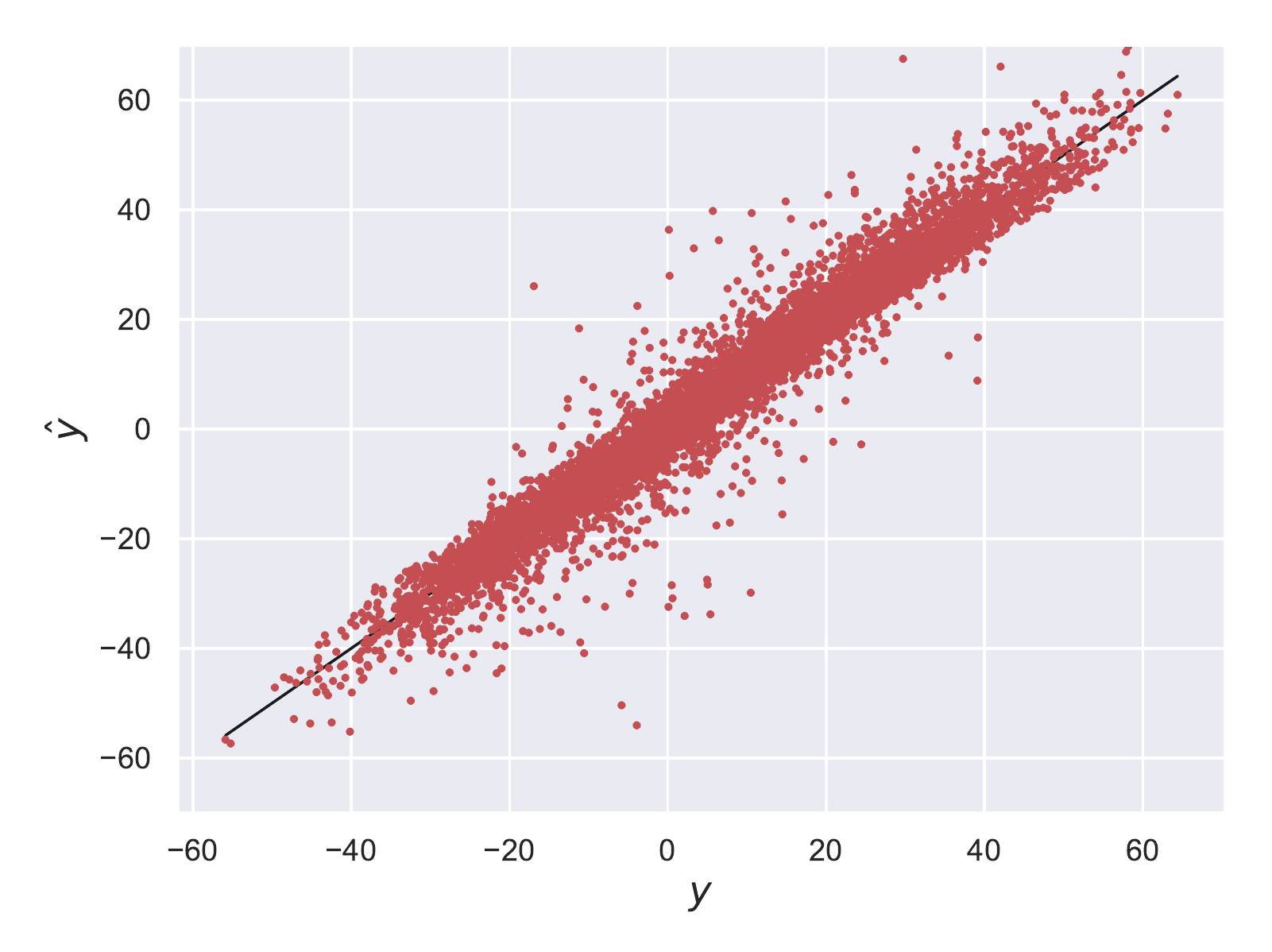}
\caption{Scatter plots of PC output obtained by CS (left) and by RVM model (right) vs true model output evaluated on the same Monte Carlo input samples. \label{fig:toy_qq}}
\end{figure}

We first study the effect of different hyperparameters $c$ in the prior Beta$(c,1)$ distribution assigned on $\bpi$, by running Algorithm \ref{alg:VI} using an ensemble of $600$ Monte Carlo samples as our data $\calD$. Table \ref{tab:toy_stats1} shows the estimates of the first 4 moments of the QoI based on direct Monte Carlo sampling from the true function and Polynomial Chaos expansions or order $P = 4$ obtained using compressive sensing and the proposed RVM with prior $c \in  \{0.2, 0.4, 0.6, 0.8,1\}$. For the interval estimates obtained using Monte Carlo with the true function, we have used $10^{5}$ samples from which we bootstrapped 1000 times and we present the $2.5$th and $97.5$th percentiles. In addition, sparsity indices are provided for the solutions that were obtained using RVM and give a percentage of the coefficients of the PCE that correspond to posterior success probabilities $> 0.01$ and $>0.95$. Fig. \ref{fig:toy_coeffs} shows the posterior means of the coefficients $\bw$ of the solution obtained for $c = 0.2$ with their $2$-standard deviation error bars, compared with those obtained using the CS algorithm (top figure) and the posterior success probabilities $\tilde{\bpi}$ of $\biota$ for the various choices of $c$ (bottom right figure). We observe good agreement between the coefficient estimates provided by the two methods, particularly the ones that appear to be far from zero and therefore correspond to the most significant terms. In areas past the $100$th coefficient, we observe also that many of the coefficients obtained from the CS method appear to be non-zero while very few of those obtained using RVM are non-zero. RVM thus provides enhanced sparsity results compared to the current CS setting. When varying $c$, we notice that the posterior estimates for $\tilde{\pi}_i$'s differ. Specifically, in the case $c = 0.2$ the posteriors success probabilities have converged to either $0$ or $1$, which makes it clear whether the corresponding term should be discarded or not, while for all other cases, the lowest values are nonzero and increase as $c$ increases to $1$. This means that sparsity in the solution is lost as $c$ increases.
This is due to the fact that the terms with corresponding success probability for $\iota_i$ being less than $1$ are still very likely to be kept in the chaos representation. The sparsity indices shown in Table \ref{tab:toy_stats1} illustrate this fact. The same $4.7\%$ of the basis terms in the $c = 0.2$ case have success probabilities that are both $>0.01$ and $>0.95$ while in the remaining cases, the terms that are ``active" with probability $>95\%$ vary from $10.3\%$ up to $15.5\%$ and all the remaining ones are active with probability between $0.08$ ($c = 0.4$) and $0.5$ ($c = 1$). We remark that the choice of $c$ with $d = 1$ implies a prior mean for the $\pi$'s equal to $c / (c+1)$. This indicates that, the smaller $c$ is selected to be, the more enhanced is the sparsity of the prior model. Additionally, the Beta density function with $c<d$ has positive skewness, meaning that a choice with $c < d$ will favor sparsity. The bottom left graph in Fig. \ref{fig:toy_coeffs} shows how the prior Beta density function $p(\pi|c, d)$ changes with $c$. Black vertical lines indicate the position of the mean for each case.

In Fig. \ref{fig:pdf_elbo} left, the probability densities of the QoI as the output of all PC expansions obtained using different $c$ values and the one obtained using CS are depicted; they all agree well. Fig. \ref{fig:pdf_elbo} right, shows plots of the ELBO values attained during optimization. As expected, the one corresponding to $c = 0.2$, that is the one with the highest sparsity and therefore fewer coefficients to be optimized, converges the fastest. As $c$ increases, convergence becomes slower and at the same time the maximum value increases as well. This is also expected, since the ELBO function involves a summation of entropy terms; as more coefficients become important, the corresponding entropy terms and those related to their hyperparameters increase their contribution. Fig. \ref{fig:toy_qq} shows scatter plots of Monte Carlo samples obtained from the two PC expansions (CS and RVM with $c = 0.2$) versus the true model output evaluated on the same input samples. The CS solution leads to a larger scatter, which is reflected in the coefficient of determination, which is $R^2 = 0.8658$ for CS and $R^2 = 0.9456$ for RVM.

\begin{figure}[h]
\centering
\includegraphics[width = 0.49\textwidth]{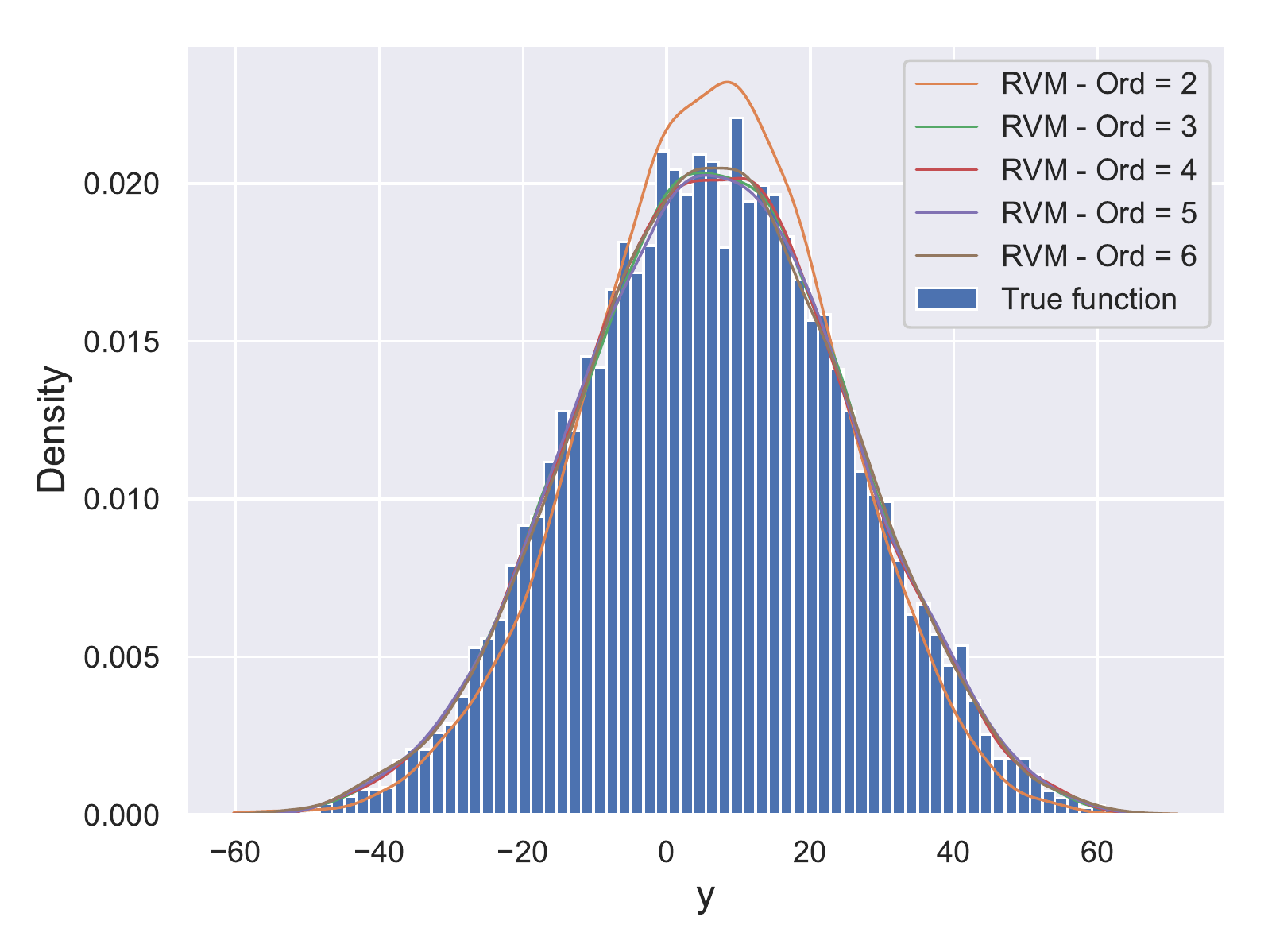}
\includegraphics[width = 0.49\textwidth]{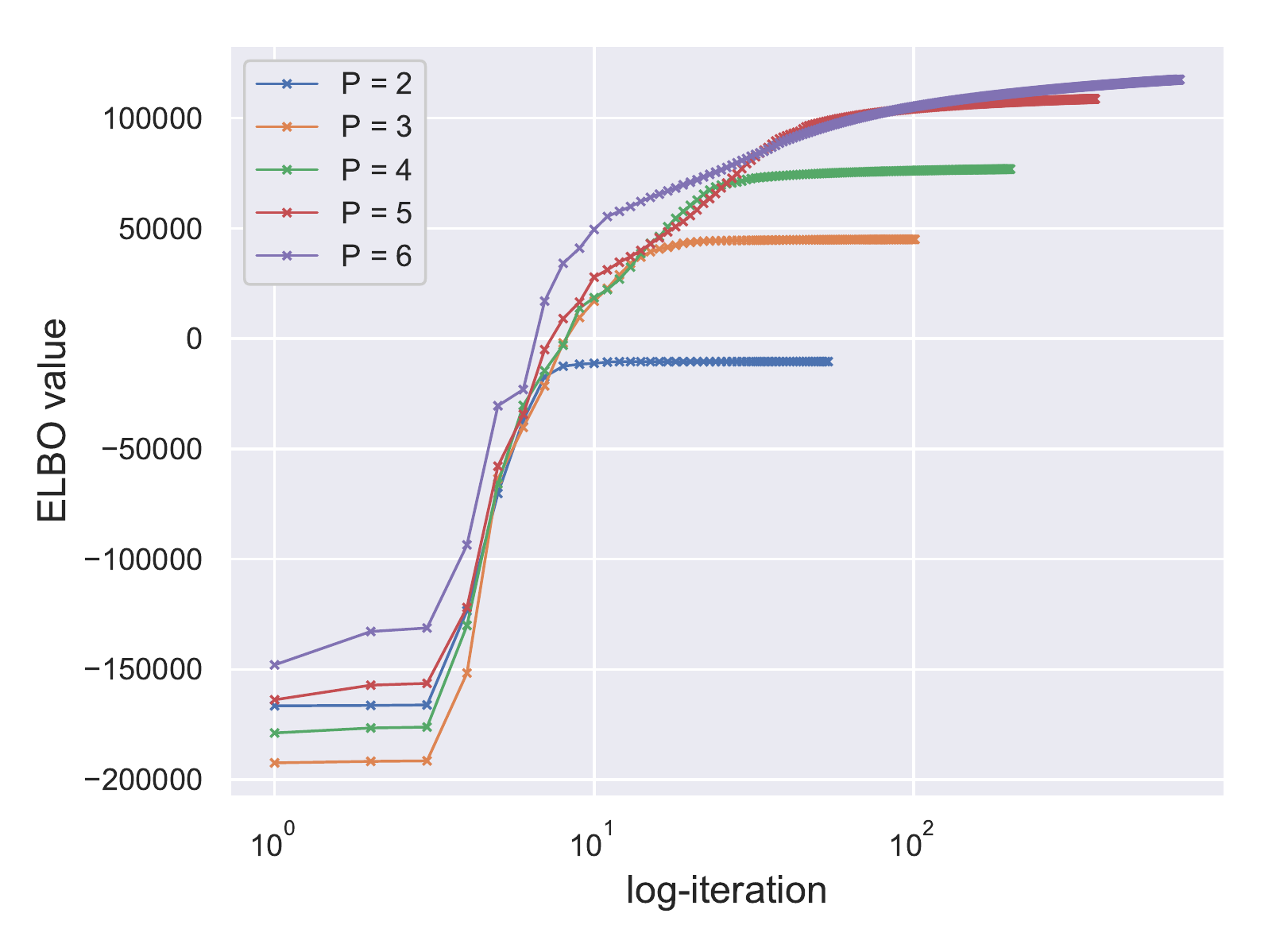}
\includegraphics[width = 0.49\textwidth]{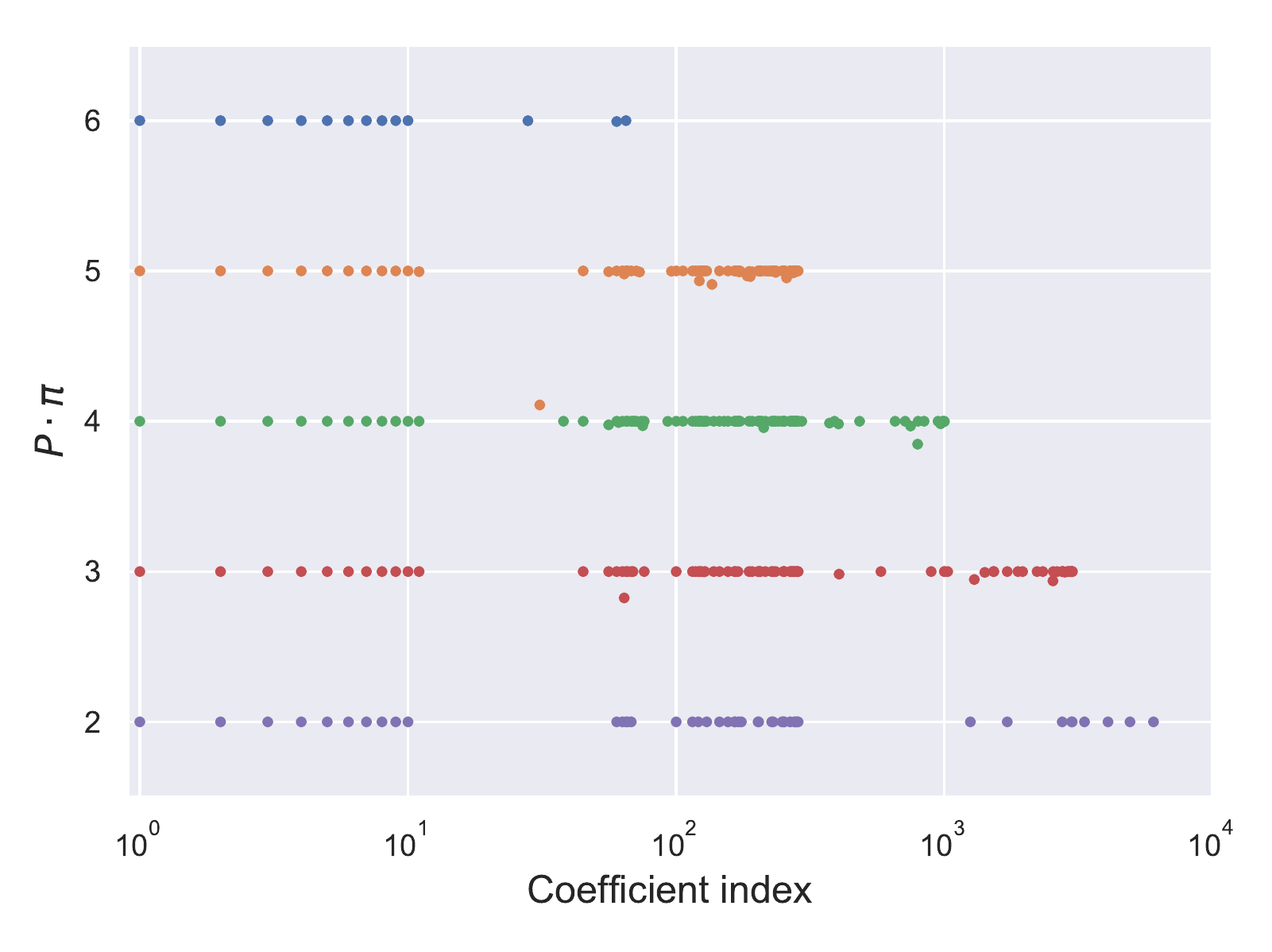}
\includegraphics[width = 0.49\textwidth]{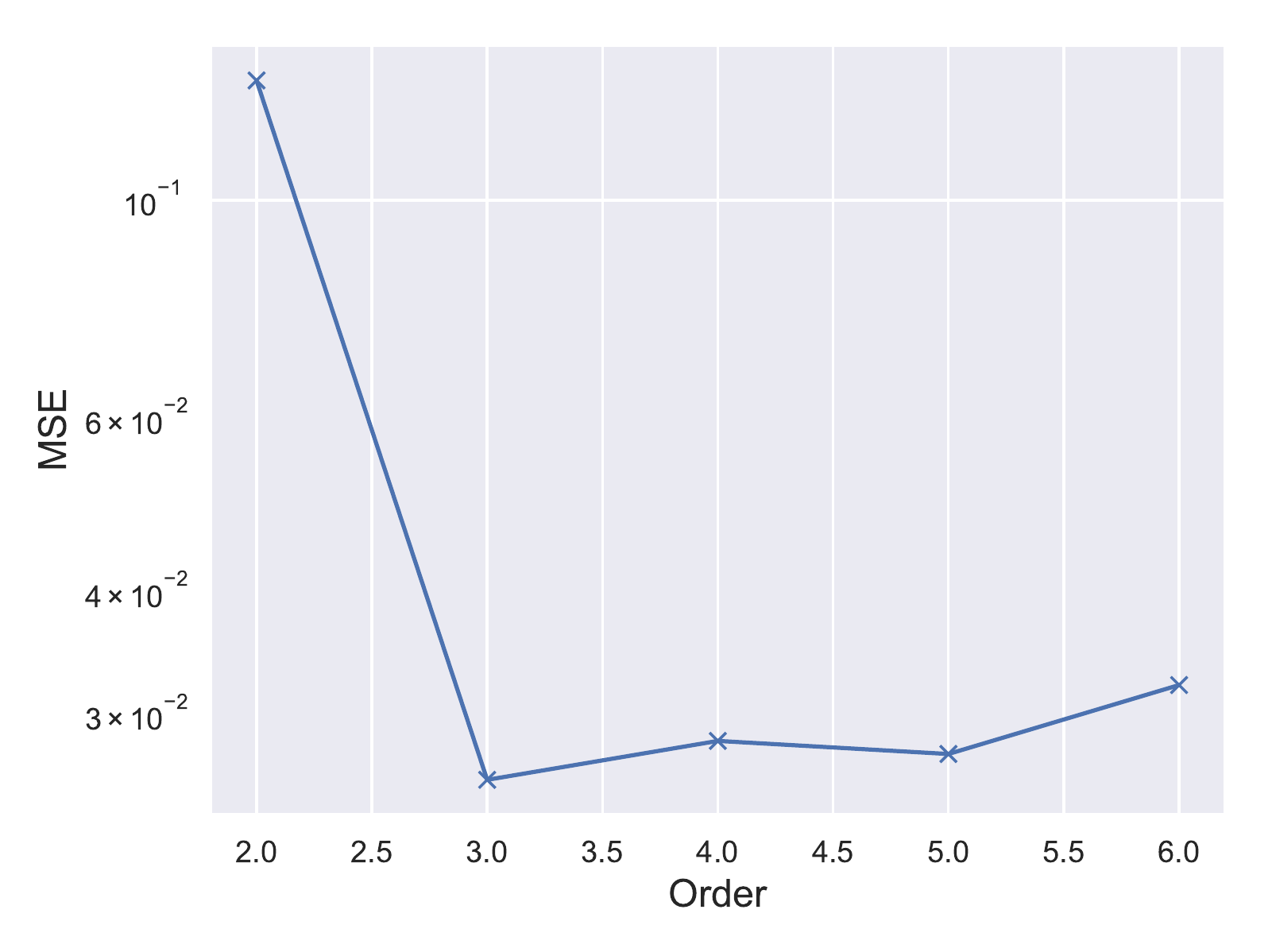}
\caption{Top left: Probability densities of the PC expansions obtained using the RVM model with varying order of truncation $P$. Top right: ELBO function evolution vs number of iterations of Algorithm \ref{alg:VI}. Bottom left: Rescaled posterior success probabilities ($P\cdot\bpi$) for PC expansions of different orders of truncation. Bottom right: Empirical relative $L_2$ of the obtained PC expansions vs truncation order. \label{fig:toy_vsord}}
\end{figure}

\begin{table}[h]
\caption{Synthetic example - Estimation of the first four statistical moments for PC expansions with varying truncation order and comparison with true model. The intervals corresponding to the true model were obtained using bootstrap resampling.}
\label{tab:toy_stats_ord}
\centering 
\begin{tabular}{l | c | r | r | r | r | r}
 & MC & $P = 2$ & $P = 3$ & $P = 4$ & $P = 5$ & $P = 6$ \\ 
 \hline 
 \hline 
 Mean & $[6.417, 6.650]$ & $6.564$ & $6.546$ & $6.561$ & $6.547$ & $6.632$ \\
 Standard deviation & $[18.59, 18.74]$ & $17.06$ & $18.80$ & $18.80$ & $18.95$ & $18.92$ \\
 Skewness & $[-0.039, -0.015]$ & $-0.102$ & $-0.037$ & $-0.021$ & $-0.020$ & $-0.030$  \\
 Kurtosis & $[2.679, 2.718]$ & $3.047$ & $2.723$ & $2.756$ & $2.764$ & $2.784$  \\
 & & & & & & \\
 Number of coefficients & & $66$ & $286$ & $1001$ & $3003$ & $8008$ \\
 Number of active coefficients & & $14$ & $81$ & $97$ & $88$ & $47$ \\
 Sparsity Index ($\tilde{\pi}_i > 0.01$) & & $21.2\%$ & $28.3\%$ & $9.7\%$ & $2.9\%$ & $0.5\%$  
\end{tabular}
\end{table}

We also test the results obtained from Algorithm \ref{alg:VI} using a PC model with varying order of truncation. Specifically, the PC expansions of order $2$ up to $6$ obtained using our method are analyzed in Fig. \ref{fig:toy_vsord}. Here we have used $1000$ Monte Carlo model evaluations as our dataset while the Beta parameter is fixed to $c = 0.2$. The top left graph shows probability densities of all PC expansions. All except the one corresponding to the lowest order of truncation ($2$), result in identical distributions. The top right graph shows the ELBO values versus the iteration steps of the optimization algorithm. When using a PC of order $2$, the algorithm converges fastest due to the smallest number of parameters to be infered, while the number of required iterations in order to achieve convergence increases gradually as the order of truncation increases. The same holds for the attained maximum value, which can also be explained by the increasing number of unknown parameters. The bottom left graph shows the rescaled values of the $\tilde{\pi}_i$ entries that have converged to $1$ after being multiplied by the corresponding order of truncation $P$. Entries with values below $0.1$ are ignored. On the x-axis is the coefficient index and on the y-axis is the truncation order of the PC expansion. Clearly the number of significant terms increases as the order increases, however, the ratio of those terms over the total number of coefficients drops. The exact values are shown as the sparsity index in Table \ref{tab:toy_stats_ord}, which drops from $21.2\%$ to only $.5\%$. At last, the bottom right graph shows the empirical relative $L_2$ error as a function of the polynomial order of truncation which remains quite stable once the polynomial order is $3$ or higher.

\subsection{Steel plate}

\begin{figure}[h]
\centering 
\includegraphics[width = 0.5\textwidth]{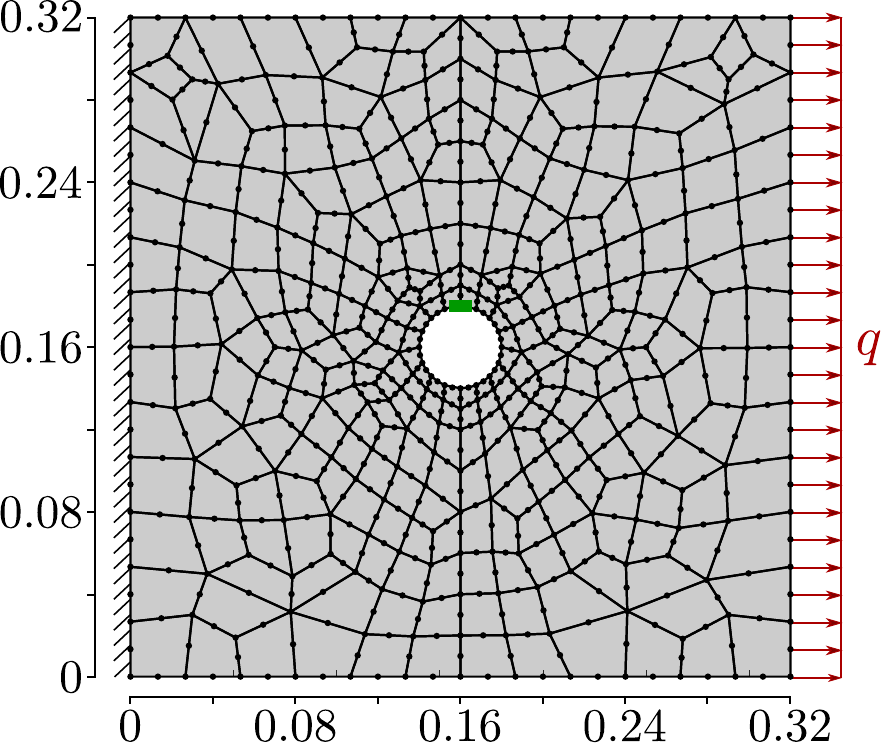}
\caption{FE-mesh of 2D-plate model. The location of maximum first principal stress $\sigma_1$ is denoted with the green marker.\label{fig:plate_mesh}}
\end{figure}

Here we consider a square low-carbon steel plate of width and length $0.32$ m, thickness $0.01$ m and a hole of radius $0.02$ m located at its center; this is a modified version of the example given in \cite{liu_and_liu}. The Poisson ratio is set to $\nu = 0.29$ and the density of the plate is $\rho = 7850$ kg/m$^3$. The horizontal and vertical displacements are constrained at the left edge. The plate is subjected to a random uniform tension $q$ that is modeled by a Gaussian $\calN(\mu_q, \sigma_q^2)$ random variable with mean $\mu_q = 60$ MPa and standard deviation $\sigma_q = 18$ MPa and is applied on the right edge. The Young's modulus $E(x,y)$ is uncertain and spatially variable. It is described by a homogeneous random field with lognormal marginal distribution with mean value $2\times10^5$ MPa and standard deviation $4\times10^4$ MPa. The autocorrelation function of the underlying Gaussian field $\ln E$ is modeled by the isotropic exponential model, $\rho_{\ln E}(\Delta x, \Delta y) = \exp\left(-\sqrt{\Delta x^2 + \Delta y^2} / l\right)$ with correlation length $l = 0.16$ m. The random field $\ln E$ is discretized by a KL expansion with $M = 37$ terms, which yields a global relative variance error of $10\%$. The stress ($\bsigma(x, y) = [\sigma_x(x,y), \sigma_y(x,y), \tau_{xy}(x, y)]^T$), strain ($\bepsilon(x,y) = [\epsilon_x(x,y), \epsilon_y(x,y), \gamma_{xy}(x, y)]^T$) and displacement ($\bu(x,y) = [u_x(x, y), u_y(x,y)]^T$) fields of the plate are given through elasticity theory, namely the Cauchy-Navier equations \cite{johnson_2009}. Given the configuration of the plate, the model can be simplified under the plane stress hypothesis, which yields
\begin{equation}\label{eq:plate}
 G(x,y)\nabla^2\bu(x,y) + \frac{E(x,y)}{2(1-\nu)} \nabla(\nabla\cdot\bu(x,y)) + \mathbf{b} =0 ~.
\end{equation}
Therein, $G(x,y) := E(x,y)/ \left(2(1 + \nu)\right)$ is the shear modulus and $\bb = [b_x, b_y]^T$ is the vector of body forces acting on the plate. Eq. \ref{eq:plate} is discretized with a finite-element method. Specifically, the spatial domain of the plate is discretized into $282$ eight-noded quadrilateral elements, as shown in Fig. \ref{fig:plate_mesh}. The scalar model output is the first principal plane stress 
\begin{equation}
\sigma_1 = 0.5\left(\sigma_x + \sigma_y\right) + \sqrt{\left[0.5(\sigma_x + \sigma_y)\right]^2 + \tau_{xy}^2}
\end{equation}
at node $11$ (see green marker in Fig. \ref{fig:plate_mesh}), which is where maximum plane stresses typically occur in this setting.

\begin{figure}[h]
\centering 
\includegraphics[width = \textwidth]{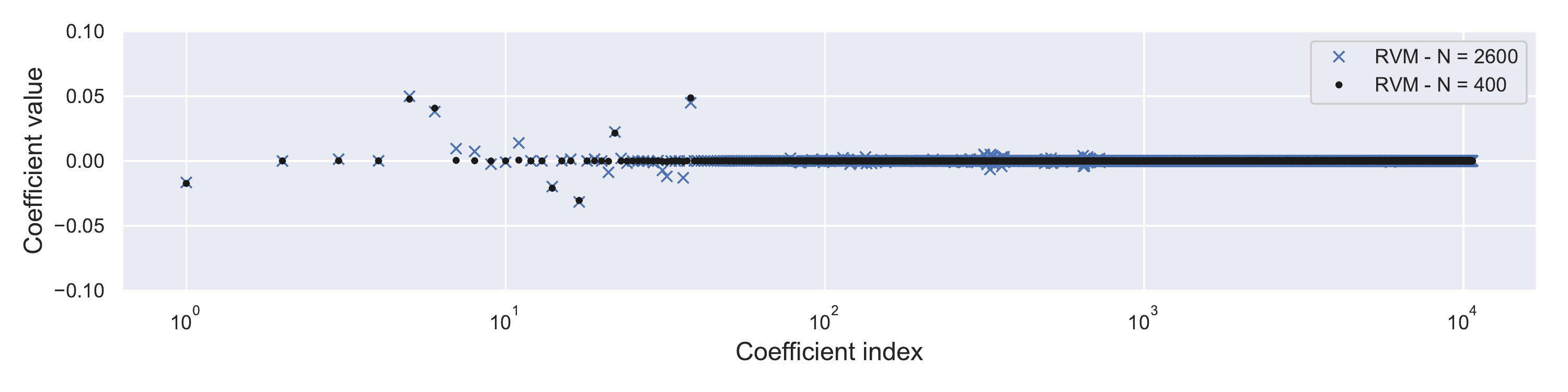}
\includegraphics[width = \textwidth]{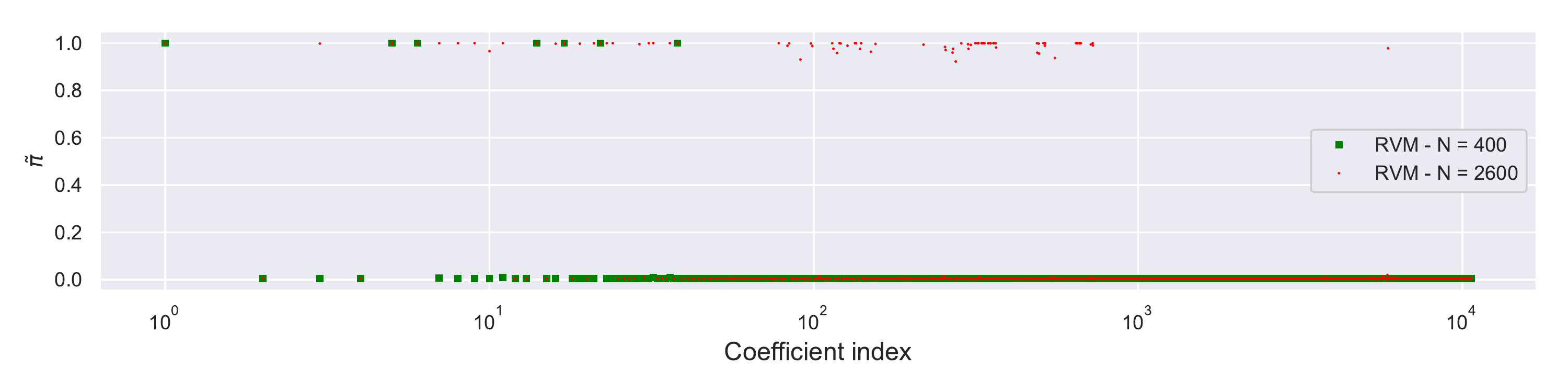}
\includegraphics[width = \textwidth]{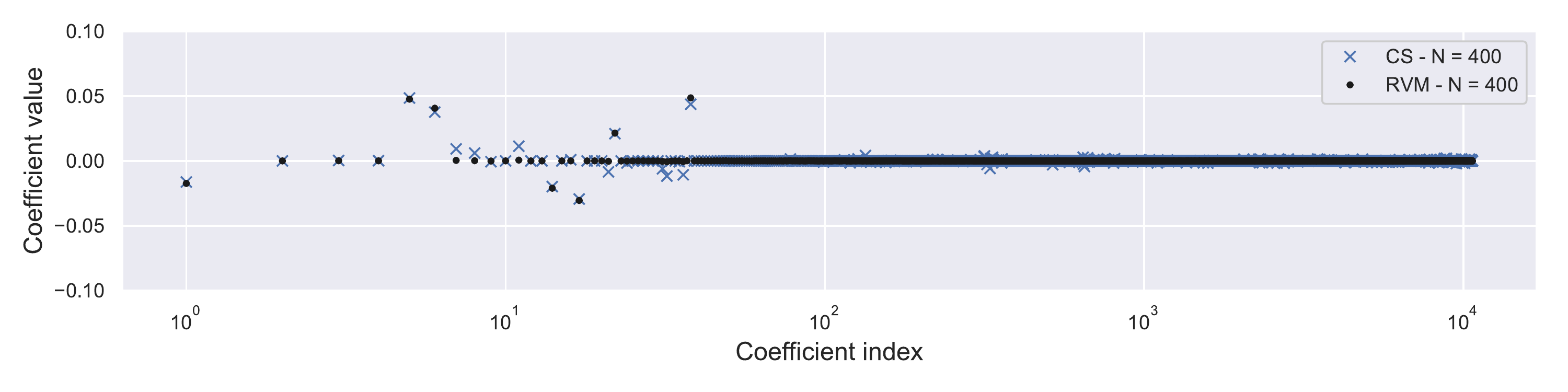}
\caption{Top: Comparison of the PC coefficients obtained using RVM with $N = 400$ and $N=2600$. Middle: The posterior success probabilities $\tilde{\bpi}$ of the PC coefficients obtained using RVM with $N=400$ and $N=2600$. Bottom: Comparison of the PC coefficients obtained using RVM and CS with $N = 400$. \label{fig:plate_coeffs}}
\end{figure}

\begin{figure}[t]
\centering
\includegraphics[width = 0.49\textwidth]{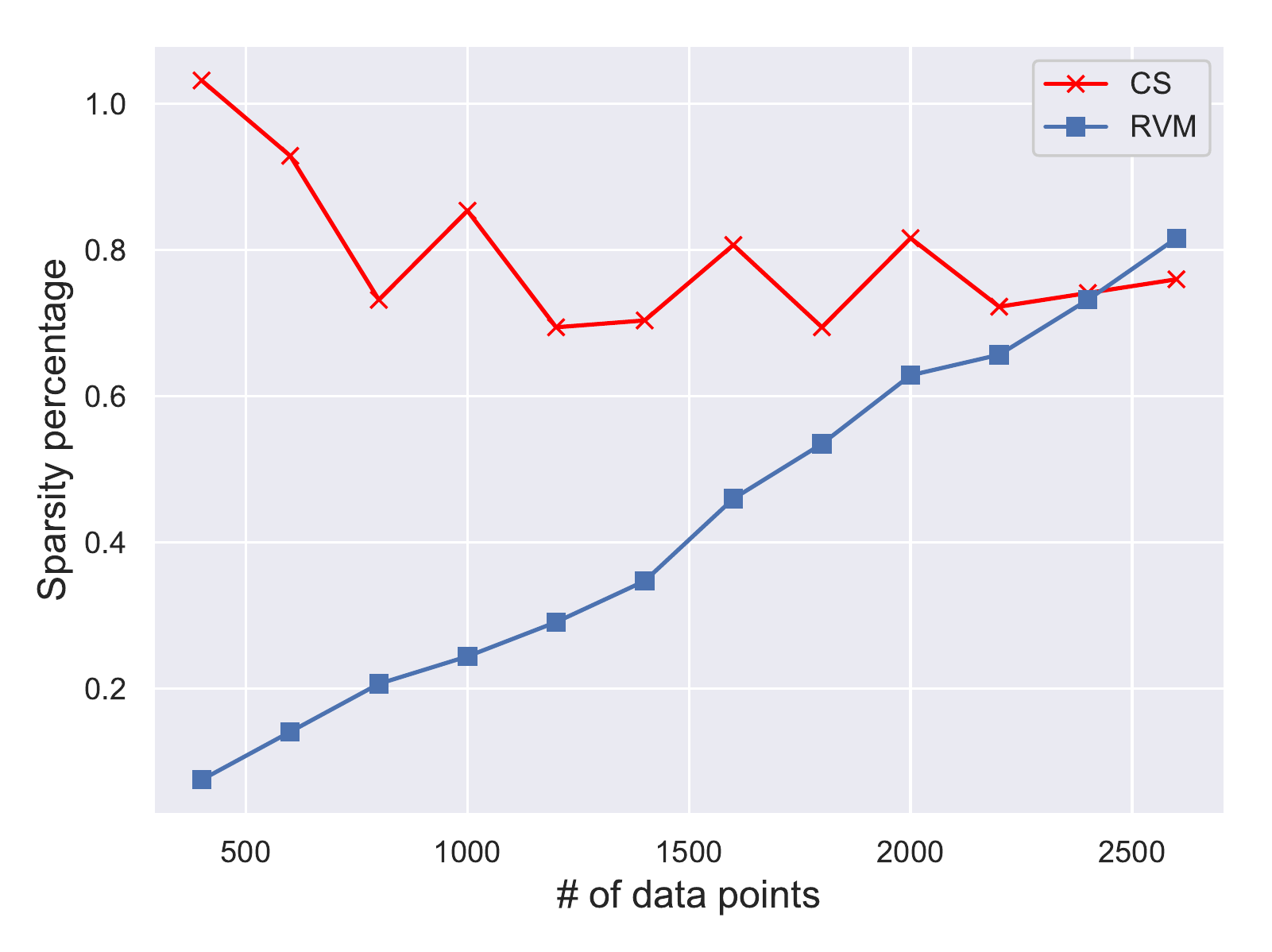}
\includegraphics[width = 0.49\textwidth]{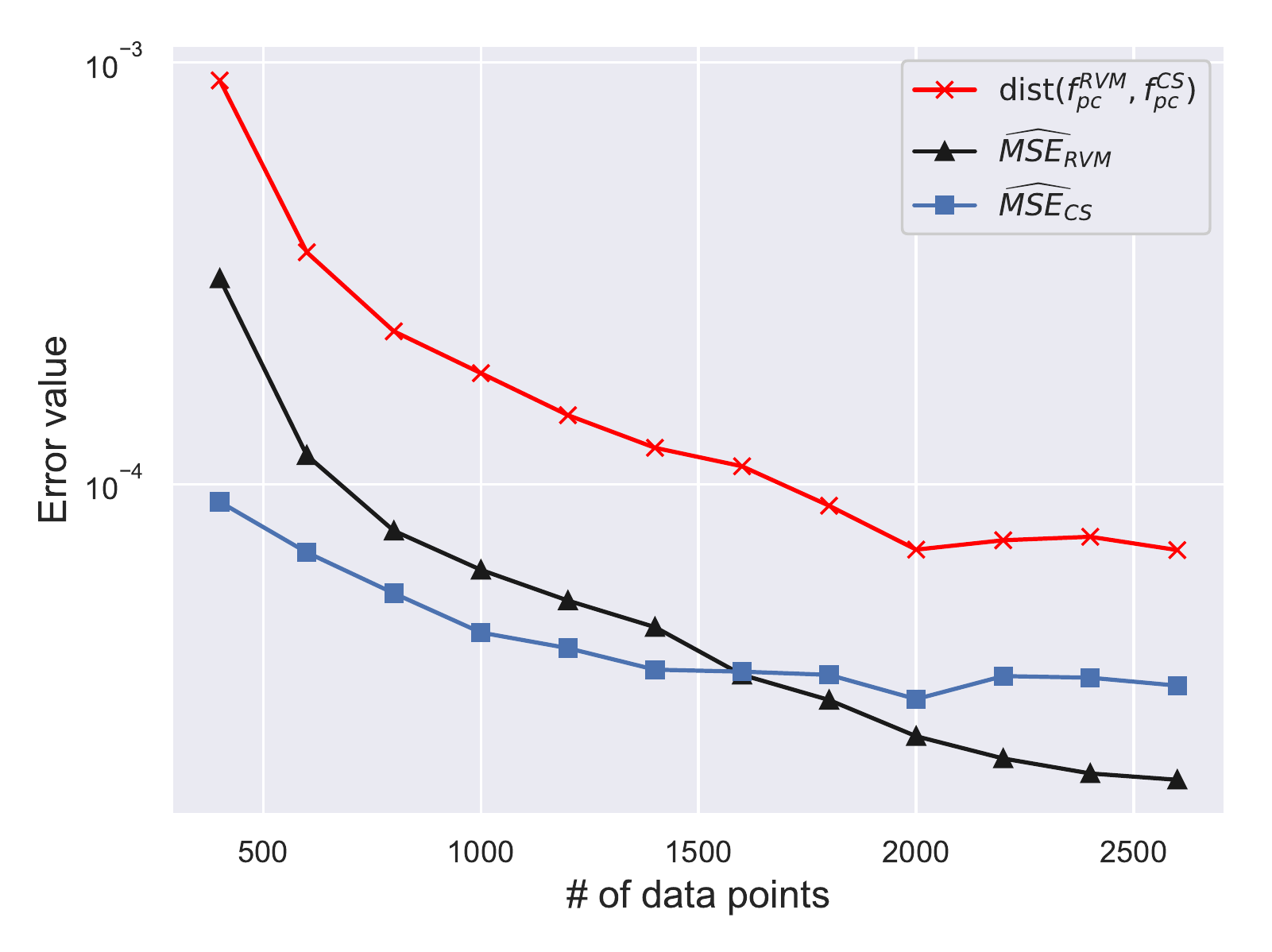}
\includegraphics[width = 0.49\textwidth]{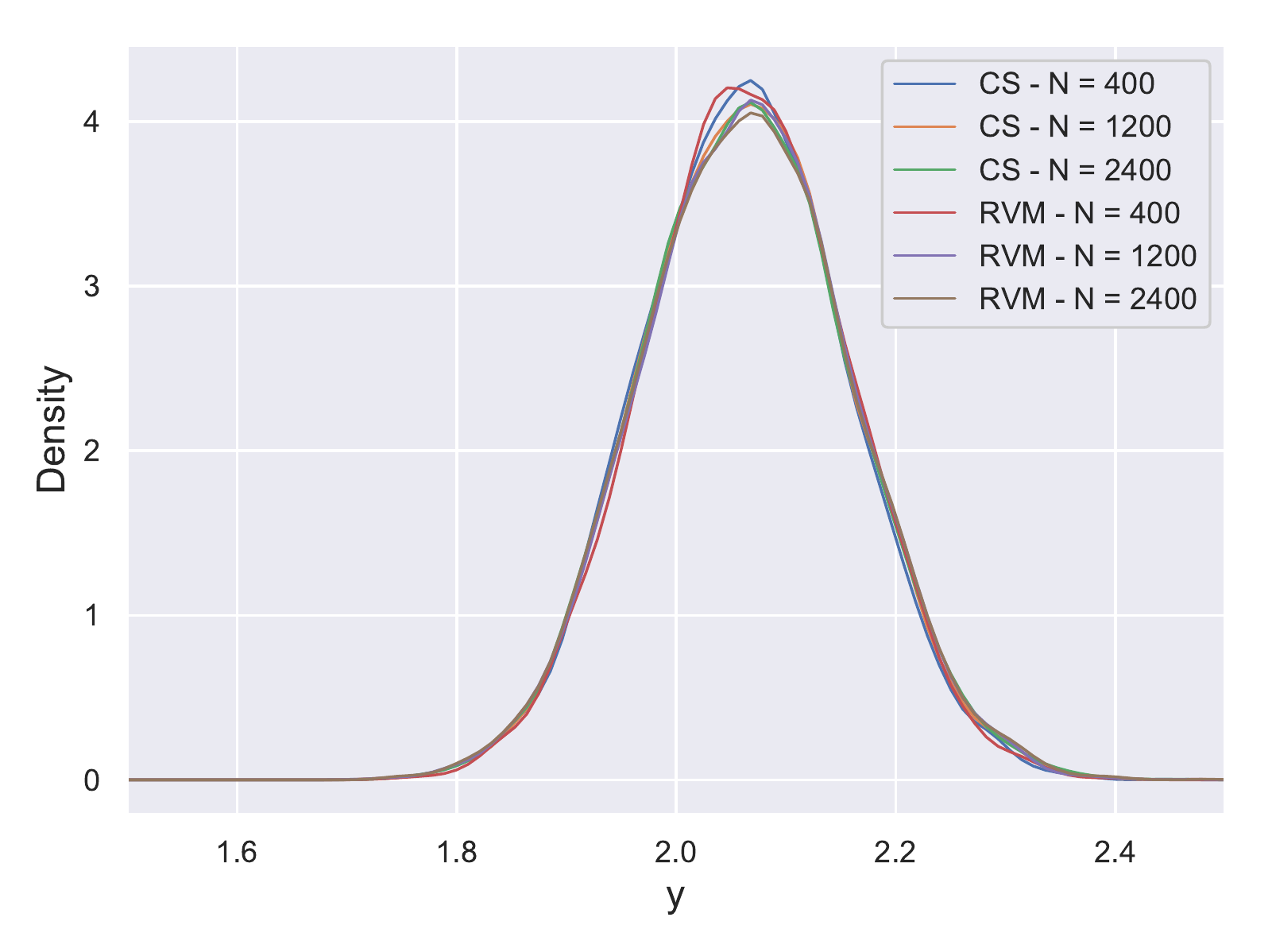}
\caption{Top left: Sparsity percentage of the PC representations obtain using CS and RVM methods, as a function of the dataset size $N$. Top right: Empirical mean square errors for CS and RVM methods and their relative MSE as a function of the dataset size $N$. Bottom: Probability density functions of the PC expansions obtained using CS and RVM using different dataset sizes $N$.\label{fig:plate_mse}}
\end{figure}

We run again Algorithm \ref{alg:VI} for a varying number of data points $N$, starting from $N = 400$ and up to $N = 2600$ and fitting a PCE of order $3$ that consists of $10660$ basis terms. For this example, we fix the Beta prior parameters to $c = 0.2$, $d = 1$. We observe that when $N$ increases, sparsity is reduced in the posterior RVM, a plausible consequence of the fact that the method attempts to fit the model on more and more data points, therefore more basis terms need to be employed in order to ``enrich" the PCE's behavior. In contrast, the PCE that we recover using the CS method does not seem to change much as a function of $N$. Fig. \ref{fig:plate_coeffs} compares the coefficients of the PCEs from the RVM method for $N = 400$ and $2600$ (top) and those obtained using RVM with $N = 400$ and CS with $N = 400$ (bottom). It is clear in both comparisons that the expansion obtained using RVM with $N = 400$ has significantly more coefficients set to zero. In fact no coefficient beyond the first $10^2$ has nonzero posterior success probability. Comparison of sparsity can also be observed in Fig. \ref{fig:plate_mse} - top left, that shows the sparsity percentage, that is the ratio of significant coefficients in the PCE's divided by the total number of coefficients, versus number of samples used in the training procedure. In the RVM method we identify the significant coefficients as in the previous example, that is the ones that have a corresponding posterior success probability $95\%$, whereas in the CS method we simply consider all coefficients with absolute values $\vert \ww_i \vert > 8\cdot10^{-4}$. It can be seen that sparsity in RVM method increases gradually from $0.07\%$ to roughly $0.8\%$, whereas in the CS case, it fluctuates between $0.7\%$ to $1\%$. In both cases, all significant coefficients correspond to polynomial basis terms of order up to $2$ (indices among the first $780$), indicating that even a second order PCE would suffice as a training model for this computational example. However, fitting a $3$rd order PC expansion with a data set as small as in this case ($N < 1000$) would be infeasible using other traditional methods such as least squares or numerical integration. Next, comparison of the empirical MSE for the two methods and the $L_2$ distance between the two PC expansions as a function of the number of samples used for training the PCE's, is shown in Fig. \ref{fig:plate_mse} (top right). For computing the $\widehat{MSE}$ we use an ensemble of $7500$ validation points. We observe that the $\widehat{MSE}$ values for the CS method are lower for small number of data points, that is, CS achieves a slightly better fit at small numbers of training points. The $\widehat{MSE}$ of the RVM improves rapidly as more data points are added and drops below that of the CS method. The $L_2$ distance between the two PCE's is not directly comparable to the empirical MSE's but simply shows that the two PCE's are getting ``closer" to each other (in the $L_2$ sense), when both algorithms use an increasing number of data points. Finally the corresponding density functions are shown in Fig. \ref{fig:plate_mse} (bottom graph). We observe that all PCE's converge in distribution and therefore, provide similar statistics.

\section{Conclusion}

We have presented a novel methodology for computing coefficients of PC expansions while accounting for sparsity in the representation.
The proposed approach provides a reduced PCE that is easy to evaluate by making use of only the terms that are marked as significant. The level of significance is quantified by assigning random sparsity indices that are multiplied with the corresponding chaos coefficients.
We model the indices as Bernoulli random variables whose success probabilities are inferred as part of the training procedure of a relevance vector machine.
We then employ variational inference to approximate the posterior distribution of the chaos coefficients and sparsity indices.
As significant terms, we eventually consider the basis terms with corresponding posterior success probability above a user-defined threshold. The proposed method is comparable to other sparse learning methods such as compressive sensing. Its performance in the numerical examples considered in this manuscript demonstrates in several occasions that it can improve sparsiy and provide a more convenient representation, compared to a standard CS algorithm. 

Several characteristics of the proposed methodology and algorithm leave room for further improvement. For instance, the optimization algorithm can be further accelerated using more efficient schemes and, perhaps, parallelization in the parameter update. Another challenge is the adaptation of the method to big data cases, using a stochastic variational inference procedure along the lines of \cite{hoffman}, where the optimization employs a Robbin-Monro scheme where at each iteration, only a subset of the available data points is used. Although these techniques were outside the scope of this paper, we consider them as promising research directions for future work on this topic.

\appendix 

\section{Exponential family distributions}
\label{sec:appA}

\subsection{Gaussian distribution}

For a Gaussian distribution with mean $m$ and scale parameter $\rho$ we have
\begin{equation}
\calN(\ww | m, \rho^{-1}) = h(\ww) \exp\left\{ \bfeta^T R(\ww) - A(\bfeta)\right\},
\end{equation}
where $\bfeta = \left(m \rho, -\rho / 2 \right)^T$, $h(\ww) = 1/ \sqrt{2\pi}$, $R(\ww) = (\ww, \ww^2)^T$ and $A(\bfeta) = -\frac{\eta_{1}^2}{4\eta_{2}} - \frac{1}{2} \log (-2\eta_2)$.

\subsection{Gamma distribution}

For the Gamma distribution with parameters $\kappa$ and $\lambda$ we have
\begin{equation}
\calG(\varsigma | \kappa, \lambda) = h(\varsigma) \exp\left\{\bfeta^T R(\varsigma) - A(\bfeta)\right\}
\end{equation}
where $\bfeta = (\kappa-1, -\lambda)^T$, $h(\varsigma) = 1$, $R(\varsigma) = (\log \varsigma, \varsigma)^T$ and $A(\bfeta) = \log\Gamma(\eta_1+1) - (\eta_1+1)\log(-\eta_2)$.

\subsection{Bernoulli distribution}

For a Bernoulli distribution with success probability $\pi$ we have 
\begin{equation}
\sfB(\iota | \pi) = h(\iota) \exp\left\{\bfeta^T R(\iota) - A(\bfeta)\right\}
\end{equation}
where $\eta = \log\left[\frac{\pi}{1-\pi}\right]$, $h(\iota) = 1$, $R(\iota) = \iota$ and $A(\eta) = \log[1 + \exp(\eta)]$.

\subsection{Beta distribution}

For a Beta distribution with parameters $r$ and $s$ we have 
\begin{equation}
\calB(\pi| r, s) = h(\pi) \exp\left\{\bfeta^T R(\pi) - A(\bfeta)\right\}
\end{equation}
where $\bfeta = (r, s)^T$, $h(\pi) = \frac{1}{\pi(1 - \pi)}$, $R(\pi) = \left(\log(\pi), \log(1 - \pi)\right)^T$ and $A(\bfeta) = \log \Gamma(\eta_1) + \log \Gamma(\eta_2) - \log \Gamma(\eta_1 + \eta_2)$.

\section{Gradients of $\calF[q]$ with respect to the natural parameters}
\label{sec:appB}

Below we compute the partial derivatives $\nabla_{\bfeta_i}\calF[q]$ analytically. First, for $\bfeta_\tau$ corresponding to $q(\tau)$ we have 
\begin{eqnarray}
\nabla_{\bfeta_\tau} \calF[q] & = & \nabla_{\bfeta_\tau}\E_q[\log p(\by
                                   | \btheta)] + \nabla_{\bfeta_\tau}
                                   \int q(\tau)\log p(\tau)d\tau -
                                   \nabla_{\bfeta_\tau} \int
                                   q(\tau)\log q(\tau) d\tau \nonumber
  \\ & = & 
\bH_\tau \E[\bL] + \bH_\tau  \bzeta_\tau - \bH_\tau \bfeta_\tau = \bH_\tau \left( \E[\bL] + \bzeta_\tau - \bfeta_\tau\right),
\end{eqnarray}
where $\bH_\tau$ is the Hessian of $A(\bfeta_\tau)$ which gives 
\begin{eqnarray}
\bH_\tau = \left[\begin{array}{cc} \phi^{(1)}(\eta_1+1) & -\frac{1}{\eta_2} \\ -\frac{1}{\eta_2} & \frac{\eta_1+1}{\eta_2^2} \end{array}\right],
\end{eqnarray}
and is nonsingular, therefore the gradient vanishes at
\begin{equation}
\bfeta_\tau = \bzeta_\tau + \E[\bL].
\end{equation}
In the above, $\phi^{(1)}(\eta) = \frac{\Gamma''(\eta)}{\Gamma(\eta)} - \frac{\Gamma'(\eta)^2}{\Gamma(\eta)^2}$ is the trigamma function $\phi^{(1)}(\eta) = \frac{d^2}{d \eta^2}\log \Gamma(\eta)$.

Second, for $\bfeta_\varsigma$ corresponding to $q(\varsigma_i)$'s, we have that the derivative of the expected log-likelihood term vanishes as it does not depend on $\varsigma_i$ therefore we get 
\begin{eqnarray}
\nabla_{\bfeta_\varsigma}\calF[q] & = & \nabla_{\bfeta_\varsigma} \E_q[\log
                                    p(\btheta)] +
                                    \nabla_{\bfeta_\varsigma}\calH[q]
                                    \nonumber \\ & = & \nabla_{\bfeta_\varsigma}\E[\bzeta_{\ww}]^T \nabla_{\bfeta_{\ww}} A_{\ww}(\bfeta_{\ww}) - \nabla_{\bfeta_\varsigma}\E[A_{\ww}(\bzeta_{\ww})] + \bH_\varsigma \bzeta_\varsigma - \bH_\varsigma \bfeta_\varsigma
\end{eqnarray}
where 
\begin{eqnarray}
\nabla_{\bfeta_\varsigma}\E[\bzeta_{\ww}]^T = -\frac{1}{2}\left[\begin{array}{cc} 0 & -\frac{1}{\eta_2} \\ 0 & \frac{\eta_1+1}{\eta^2_2} \end{array}\right]
\end{eqnarray}
and 
\begin{eqnarray}
\nabla_{\bfeta_\omega}\E[A_{\ww}(\bzeta_{\ww})] = -\frac{1}{2}\left[\begin{array}{c}  \phi^{(1)}(\eta_1+1) \\ -\frac{1}{\eta_2}\end{array}\right]
\end{eqnarray}
and we can write
\begin{eqnarray}
\nabla_{\bfeta_\varsigma}\calF[q] & = & \bH_{\varsigma} \left( \frac{1}{2} \left[\begin{array}{c}1 \\ -\frac{\partial}{\partial\eta_{\ww,2}}A_{\ww}(\bfeta_{\ww})  \end{array}\right] + \bzeta_\varsigma - \bfeta_\varsigma \right)
\end{eqnarray}
where $\bH_\varsigma$ has the same expression as $\bH_\tau$ and therefore the gradient vanishes at 
\begin{eqnarray}
\bfeta_\varsigma = \bzeta_\varsigma + \frac{1}{2} \left[\begin{array}{c}1 \\ - \frac{\partial}{\partial\eta_{\ww,2}}A_{\ww}(\bfeta_{\ww})  \end{array}\right].
\end{eqnarray}

Similarly, for $\bfeta_{\pi}$ that correspond to $q_{\pi_i\vert r_i, s_i}(\pi_i)$'s, we have that the expected log-likelihood derivative term also vanishes and we get 
\begin{eqnarray}
\nabla_{\bfeta_\pi}\calF[q] & = & \nabla_{\bfeta_\pi} \E_q[\log
                                    p(\bTheta)] +
                                    \nabla_{\bfeta_\pi}\calH[q]
                                    \nonumber \\ & = & \nabla_{\bfeta_\pi}\E[\bzeta_{\iota}] \nabla_{\bfeta_{\iota}} A_{\iota}(\bfeta_{\iota}) - \nabla_{\bfeta_\pi}\E[A_{\iota}(\bzeta_{\iota})] + \bH_\pi \bzeta_\pi - \bH_\pi \bfeta_\pi,
\end{eqnarray}
where $\bH_\pi$ is the Hessian of $A(\bfeta_\pi)$ which gives 
\begin{eqnarray}
\bH_\pi = \left[\begin{array}{cc}  \phi^{(1)}(\eta_1) - \phi^{(1)}(\eta_1+\eta_2) & - \phi^{(1)}(\eta_1+\eta_2) \\  - \phi^{(1)}(\eta_1+\eta_2) &  \phi^{(1)}(\eta_2) - \phi^{(1)}(\eta_1+\eta_2)\end{array}\right].
\end{eqnarray}
Moreover, we write $\nabla_{\bfeta_\pi}\E[\bzeta_{\iota}] = \bH_{\pi}\left[\begin{array}{c} 1 \\ -1 \end{array}\right]$ and $\nabla_{\bfeta_\pi}\E[A_{\iota}(\bzeta_{\iota})] = \bH_{\pi}\left[\begin{array}{c} 0 \\ -1 \end{array}\right]$
which implies 
\begin{eqnarray}
\nabla_{\bfeta_\pi}\calF[q] & = & \bH_{\pi} \left( \left[\begin{array}{c} 1 \\ -1 \end{array}\right] \nabla_{\bfeta_{\iota}} A_{\iota}(\bfeta_{\iota}) - \left[\begin{array}{c} 0 \\ -1 \end{array}\right] + \bzeta_\pi - \bfeta_\pi\right)
\end{eqnarray}
and the gradient vanishes at 
\begin{equation}
\bfeta_\pi = \bzeta_\pi + \left[\begin{array}{c} 0 \\ 1 \end{array}\right] - \left[\begin{array}{c} 1 \\ -1 \end{array}\right] \nabla_{\bfeta_{\iota}} A_{\iota}(\bfeta_{\iota}).
\end{equation}

Next, for $\eta_{\iota}$ corresponding to $q(\iota_i)$'s, we have 
\begin{eqnarray}
\frac{\partial}{\partial\eta_{\iota}}\calF[q] & = & \frac{\partial}{\partial\eta_{\iota}}\E_q[\log p(\by |
                               \btheta)] + \frac{\partial}{\partial\eta_{\iota}} \E_q[\log
                               p(\btheta)] + \frac{\partial}{\partial\eta_{\iota}} \calH[q]
                               \nonumber \\ & = & 
                               \frac{\partial}{\partial\eta_{\iota}}\E_q[\log p(\by | \btheta)] + \bh_{\iota}\left(\E[\zeta_{\iota}] - \eta_{\iota}\right),
\end{eqnarray}
where $\bh_{\iota} = A''(\eta_P{\iota}) = \frac{e^{\eta_{\iota}}}{1 + e^{\eta_{\iota}}} - \frac{e^{2\eta_{\iota}}}{\left(1 + e^{\eta_{\iota}}\right)^2}$ and 
\begin{eqnarray}
\frac{\partial}{\partial\eta_{\iota}}\E_q[\log p(\by | \btheta)] & = &
                                                  \frac{\partial}{\partial\eta_{\iota}}\left(\int q(\iota)\bL d\iota\right)^T \nabla_{\bfeta_\tau}A(\bfeta_\tau)\nonumber \\ & = &
\left[\begin{array}{cc} 0 & \frac{\partial}{\partial\eta_{\iota}} \E[L_2]\end{array}\right]\nabla_{\bfeta_\tau}A(\bfeta_\tau) = \frac{\partial}{\partial\eta_{\iota}} \E[L_2]\cdot \frac{\partial A(\bfeta_\tau)}{\partial \eta_{\tau,2}}.
\end{eqnarray}
We write $\E[L_2] = -\frac{1}{2}\big\vert \big\vert \by - \bPsi(\bpi \circ \bm)\big\vert \big\vert^2 - \frac{1}{2}\trace{\bPsi^T\bPsi\left(\diag{\bpi \circ \brho^{-1}} + \diag{\left((\mathbf{1}-\bpi) \circ \bm) \circ (\bpi\circ\bm\right)}\right)}$ and for $\bpi = [\pi_1, \dots, \pi_{N_{d'}}]^T$ with $\pi_i = e^{\eta_{\iota}} / (1 + e^{\eta_{\iota}})$, we get 
\begin{eqnarray}
\frac{\partial}{\partial\eta_{\iota}} \E[L_2] & = & \bh_{\iota} \bu
\end{eqnarray}
with
\begin{eqnarray}\bu & = & \left( \by^T \bPsi(\bepsilon_i \circ \bm) - (\bepsilon_i \circ \bm)^T \bPsi^T \bPsi (\bpi \circ \bm) \right. \\ & - & \left. \frac{1}{2}\trace{\bPsi^T\bPsi\left( \diag{\bepsilon_i \circ \brho^{-1}} + \diag{ ((\mathbf{1} - 2\bpi) \circ \bm) \circ (\bepsilon_i\circ\bm) } \right)} \right) ,
\end{eqnarray}
where $\bepsilon_i$ is the unit vector with $1$ at the $i$th position and zero elsewhere, and the gradient vanishes when
\begin{eqnarray}
\eta_{\iota} & = &  \E[\zeta_{\iota}] + \bu \frac{\partial A(\bfeta_\tau)}{\partial \eta_{\tau,2}}. 
\end{eqnarray}

At last, for $\bfeta_\ww$ corresponding to $q_{\ww_i\vert \varsigma_i}(\ww_i)$, we get 
\begin{eqnarray}
\nabla_{\bfeta_{\ww}}\calF[q] & = & \nabla_{\bfeta_{\ww}}\E_q[\log p(\by |
                               \btheta)] + \nabla_{\bfeta_\ww} \E_q[\log
                               p(\btheta)] + \nabla_{\bfeta_\ww} \calH[q]
                               \nonumber \\ & = & 
\nabla_{\bfeta_\ww}\E_q[\log p(\by | \btheta)] + \bH_{\ww}\left(\E[\bzeta_{\ww}] - \bfeta_{\ww}\right)
\end{eqnarray}
where 
\begin{eqnarray}
\bH_\ww = \left[\begin{array}{cc} -\frac{1}{2\eta_2} & \frac{\eta_1}{2\eta_2^2} \\ \frac{\eta_1}{2\eta_2^2} & -\frac{\eta_1^2}{2\eta_2^3} + \frac{1}{2\eta_2^2} \end{array}\right]
\end{eqnarray}
and
\begin{eqnarray}
\nabla_{\bfeta_{\ww}}\E_q[\log p(\by | \btheta)] & = &
\left[\begin{array}{cc} \mathbf{0} & \nabla_{\bfeta_\ww} \E[L_2]\end{array}\right]\nabla_{\bfeta_\tau}A(\bfeta_\tau).
\end{eqnarray}
For $\bfeta_\ww = (\eta_{\ww,1}, \eta_{\ww,2})$, where $\bm= [m_1, \dots, m_{N_{K}}]^T$, $\brho = [\rho_1, \dots, \rho_{N_{K}}]^T$ and $m_i = -\frac{\eta_{\ww_i,1}}{2\eta_{\ww_i,2}}$, $\rho_i = -2\eta_{\ww_i, 2}$, 
after some tedious algebraic manipulations one gets 
\begin{eqnarray}
\nabla_{\bfeta_\ww}\E_q[\log p(\by | \btheta)] & = & \nabla_{\bfeta_\ww}
                                                  \E[L_2]
                                                  \frac{\partial
                                                  A(\bfeta_\tau)}{\partial
                                                  \eta_{\tau,2}}\nonumber
  \\ & = & \bH_{\ww} \bv \cdot \frac{\partial A(\bfeta_\tau)}{\partial \eta_{\tau,2}},
\end{eqnarray}
where 
\begin{eqnarray}
\bv = \left[ \begin{array}{c}\by^T \bPsi (\bepsilon_i\circ \bpi) - (\bepsilon_i \circ \bpi )^T \bPsi^T\bPsi (\bpi \circ \bm_{-i}) \\ -\frac{1}{2}\trace{\bPsi^T\bPsi\diag{\bepsilon_i\circ\bpi} }\end{array} \right]
\end{eqnarray}
and $\bm_{-i}$ is $\bm$ with $0$ at it's $i$th entry. Finally we have 
\begin{equation}
\nabla_{\bfeta_\ww}\calF[q] = \bH_\ww \left(\bv\frac{\partial A(\bfeta_\tau)}{\partial \eta_{\tau,2}} + \E[\bzeta_\ww] - \bfeta_\ww\right)
\end{equation}
which vanishes at 
\begin{equation}
\bfeta_\ww = \bv\frac{\partial A(\bfeta_\tau)}{\partial \eta_{\tau,2}} + \E[\bzeta_\ww]. 
\end{equation}

\section*{}
\bibliographystyle{plain}
\bibliography{references}

\end{document}